\documentclass{article}

\PassOptionsToPackage{numbers, sort&compress}{natbib}

\usepackage[final]{neurips_2024}

\usepackage[utf8]{inputenc} % allow utf-8 input
\usepackage[T1]{fontenc}    % use 8-bit T1 fonts
\usepackage{url}            % simple URL typesetting
\usepackage{booktabs}       % professional-quality tables
\usepackage{amsfonts}       % blackboard math symbols
\usepackage{nicefrac}       % compact symbols for 1/2, etc.
\usepackage{microtype}      % microtypography
\usepackage{xcolor}         % colors
\usepackage{graphicx}
\usepackage[accsupp]{axessibility}  % Improves PDF readability for those with disabilities.

\usepackage{graphicx}
\usepackage{amsmath}
\usepackage{amssymb}
\usepackage{bm}
\usepackage{booktabs}
\usepackage{epsfig}
\usepackage{epigraph}
\usepackage{graphicx}
\usepackage{amsmath}
\usepackage{caption}
\usepackage[table]{xcolor}
\usepackage{threeparttable}
\usepackage{tabularx}
\usepackage{tabularray}
\usepackage{wrapfig}
\usepackage{multirow}
\usepackage{float}
\usepackage{overpic}
\usepackage{appendix}
\usepackage{float}
\usepackage{booktabs} % For formal tables
\usepackage{diagbox}
\usepackage{makecell}
\usepackage{bbm}
\usepackage{threeparttable}
\usepackage{dsfont}
\usepackage{subfigure} 
\usepackage{arydshln}
\usepackage{overpic}
\usepackage[misc]{ifsym}
\usepackage{wrapfig}
\usepackage{xspace}
\usepackage{varwidth}

\usepackage[pagebackref,breaklinks,colorlinks,citecolor=teal]{hyperref}
\usepackage{orcidlink}
\usepackage{varwidth}

\title{InterDreamer: Zero-Shot Text to 3D Dynamic Human-Object Interaction}

\author{Sirui Xu$^{\dag}$ \quad Ziyin Wang$^{\dag}$ \quad
Yu-Xiong Wang$^{\ddag}$ \quad
Liang-Yan Gui$^{\ddag}$\\
University of Illinois Urbana-Champaign\\
$^{\dag}$ Equal Contribution   \quad  $^{\ddag}$ Equal Advising\\
{\tt\small \{siruixu2, ziyin, yxw, lgui\}@illinois.edu}\\
\url{https://sirui-xu.github.io/InterDreamer/}}
\def\ours{InterDreamer}
\def\high{planning}
\def\low{control}

\def\@onedot{\ifx\@let@token.\else.\null\fi\xspace}

\begin{document}
\maketitle

\begin{abstract}
Text-conditioned human motion generation has experienced significant advancements with diffusion models trained on extensive motion capture data and corresponding textual annotations. However, extending such success to 3D dynamic human-object interaction (HOI) generation faces notable challenges, primarily due to the lack of large-scale interaction data and comprehensive descriptions that align with these interactions. This paper takes the initiative and showcases the potential of generating human-object interactions \textit{without direct training on text-interaction pair data}. Our \textit{key insight} in achieving this is that interaction semantics and dynamics can be decoupled. Being unable to learn interaction semantics through supervised training, we instead leverage pre-trained large models, synergizing knowledge from a large language model and a text-to-motion model. While such knowledge offers high-level control over interaction semantics, it cannot grasp the intricacies of low-level interaction dynamics. To overcome this issue, we introduce a world model designed to comprehend simple physics, modeling how human actions influence object motion. By integrating these components, our novel framework, InterDreamer, is able to generate text-aligned 3D HOI sequences without relying on paired text-interaction data. We apply InterDreamer to the BEHAVE, OMOMO, and CHAIRS datasets, and our comprehensive experimental analysis demonstrates its capability to generate realistic and coherent interaction sequences that seamlessly align with the text directives.
\end{abstract}
  
\section{Introduction}
\label{sec:intro}
\begin{figure*}\centering
\includegraphics[width=\textwidth]{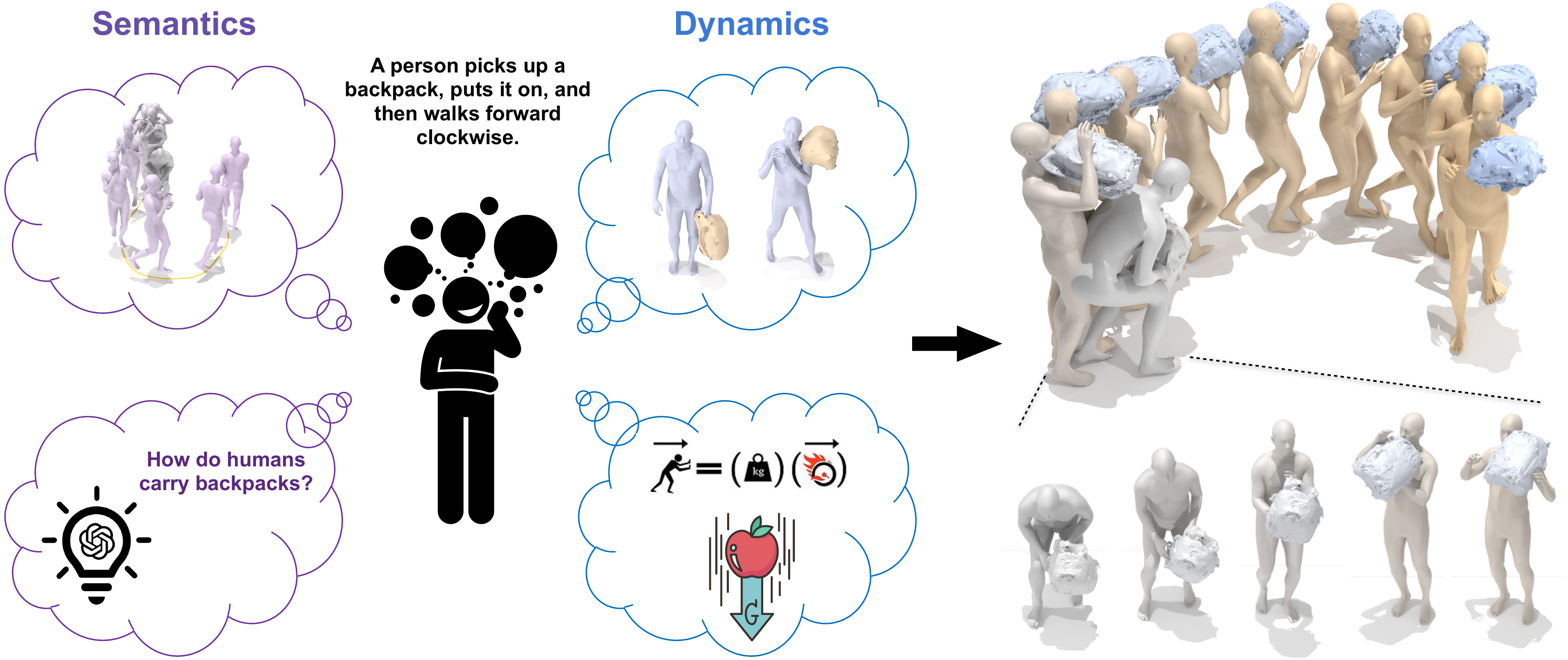}
\captionof{figure}{\ours~generates vivid 3D human-object interaction sequences guided by text descriptions, by synergizing semantics and dynamics knowledge from large-scale text-motion data ({\textcolor{purple}{upper left}}), a large language model ({\textcolor{purple}{bottom left}}), human-object interaction data ({\textcolor{blue}{upper middle}}), and prior knowledge ({\textcolor{blue}{bottom middle}}) from simple physics. We visualize the generated text-guided interaction sequence ({\textcolor{orange}{upper right}}), with the beginning of the sequence unfolded ({\textcolor{orange}{bottom right}}).
\label{fig:teaser}}
\end{figure*}

Text-guided human motion generation has made unprecedented progress through advancements in diffusion models~\cite{sohl2015deep, song2020denoising, ho2020denoising,wang2024diffusion}, leading to synthesis outcomes that are realistic, diverse, and controllable. This progress has ignited an increased interest in exploring expanded tasks related to text-guided human interaction generation, such as social interaction~\cite{liang2023intergen} and human-scene interaction~\cite{huang2023diffusion}. However, many of these explorations are limited in that the dynamics of objects is not involved or text-guided. Aiming to bridge such a gap, this paper tackles a more challenging task -- \emph{generating versatile 3D human-object interactions (HOIs) through language guidance}, as illustrated in Figure~\ref{fig:teaser}.

Although a direct solution, as suggested by the concurrent work~\cite{peng2023hoi, diller2023cg, li2023controllable, wu2024thor, wu2024human, song2024hoianimator}, would be replicating the success observed in human motion generation and adopting a similar supervised approach for learning text-driven HOIs, it is not scalable. As can be observed, generating social or scene interactions is heavily dependent on extensive collections of text-interaction pair data~\cite{mahmood2019amass, guo2022generating, wang2022humanise, liang2023intergen}, and scaling these methods to address more complex HOIs outlined in our study could require datasets of comparable magnitude. Achieving this goal appears unattainable by merely annotating existing 3D HOI datasets~\cite{bhatnagar22behave, jiang2022chairs, huang2022intercap, zhang2023neuraldome, fan2023arctic, li2023object,zhao2023im, kim2024parahome,zhang2024force,zhang2024hoi}, which are relatively limited in size. Although recent studies~\cite{peng2023hoi, diller2023cg, li2023controllable, yang2024f} have annotated some of these datasets, the volume of text-interaction pairs still lags behind that available for existing text-driven motion generation efforts.

An intriguing question naturally arises: given the limited annotations of the text, \emph{what is the potential of learning for text-conditioned HOI generation without text supervision}, which is the main focus of this paper. However, formulating the task in such a setting presents significant challenges, primarily due to the inability to directly learn the alignment between text and HOI dynamics. Our key observation is that interaction semantics and dynamics can be \textit{decoupled}. That is, the high-level semantics of an interaction, aligned with its textual description, can be informed by \textit{human motion} and the \textit{initial object pose}. Meanwhile, the low-level dynamics of the interaction -- specifically, the \textit{subsequent} behavior of the object -- is governed by the forces exerted by the human, within the constraints of physical laws. 
Motivated by these insights, we introduce InterDreamer -- a novel framework that synergizes knowledge of interaction semantics and dynamics (Figure~\ref{fig:teaser}), both of which do not necessarily require learning from text-interaction pairs, if they are decoupled.

The semantics of interaction, although not available through direct supervised training, can be harnessed from \textit{prior knowledge} without text-interaction pair datasets. 
Specifically, to acquire semantically aligned interaction, we first consult a large language model (LLM), such as GPT-4~\cite{chatgpt} and Llama 2~\cite{touvron2023llama}, to provide understanding including how humans typically use specific body parts in interactions with particular objects, by exploiting its \textit{in-context learning} capability with \textit{few-shot prompting}~\cite{brown2020language} and \textit{chain-of-thought prompting}~\cite{wei2022chain}. The intermediate thoughts and the final thought are then used to (\textbf{i}) generate semantically aligned human motion with a pre-trained text-to-motion model; and (\textbf{ii}) identify an initial object pose that is harmonious with the generated human pose and text description, following a philosophy similar to \textit{retrieval-augmented generation}~\cite{lewis2020retrieval}.

While these large models can offer high-level motion semantic modeling, they lack crucial \textit{low-level} dynamics knowledge. Nevertheless, by decoupling interaction dynamics from semantics, a key advantage emerges in our InterDreamer framework: interaction dynamics can be learned from motion capture data \textit{without the necessity of text annotations}.  
We instantiate this idea by developing a \emph{world model}, which predicts the subsequent state of an object affected by the interaction.
The key here is to reach \emph{generalizable representations} in different motion and objects.
To do so, we exert control over the object through the motion of vertices on the human body. These vertices are solely sampled in regions where contact occurs, \textit{agnostic} to the overall object shape and whole-body motion. Such abstraction empowers the model to learn the simple dynamics from a publicly available 3D HOI dataset BEHAVE~\cite{bhatnagar22behave}, and generalize naturally to other datasets~\cite{li2023object,jiang2022chairs}.
The plausibility of the generated interaction is further enhanced by a subsequent optimization procedure on the synthesized human and object motion.

To summarize, our contributions are:
(\textbf{i}) We address the task of synthesizing whole-body interactions with dynamic objects guided by textual commands, achieving this notably without the need for paired text-interaction data, a novel paradigm to the best of our knowledge. 
(\textbf{ii}) We introduce a framework that decomposes semantics and dynamics, and they can be easily integrated. Specifically, it harnesses knowledge from a large language model (LLM) and a text-to-motion model as external resources, alongside our proposed world model. Remarkably, the only component that requires additional training is the world model, which highlights the \textit{ ease of use} of our framework.
(\textbf{iii}) Experimental results demonstrate that our framework, InterDreamer, is capable of producing semantically aligned and realistic human-object interactions, and generalizes \textit{beyond existing HOI datasets}. 
\section{Related Work}
\label{sec:related}
\noindent{\bf Text-Conditioned Human Motion Generation.}
Significant progress has been witnessed in human motion synthesis tasks, given different kinds of external conditions, including action categories~\cite{guo2020action2motion,petrovich2021action,lee2023multiact,athanasiou2022teach}, past motion~\cite{yuan2020dlow, mao2021generating, barquero2022belfusion, chen2023humanmac, xu22stars, xu2023stochastic, sun2023towards}, trajectories~\cite{kaufmann2020convolutional,karunratanakul2023gmd,rempe2023trace,xie2023omnicontrol,wan2023tlcontrol,feng2024stratified}, scene context~\cite{cao2020long,hassan2021populating,wang2021synthesizing,wang2021scene,wang2022towards,wang2022humanise,huang2023diffusion,zhao2022compositional,Zhao:ICCV:2023,tendulkar2022flex,zhang2023roam,diller2024futurehuman3d,tang2024unified,xue2024shape}, and without condition~\cite{raab2023modi}.
Recently, human motion synthesis guided by textual descriptions~\cite{petrovich2023tmr,guo2022tm2t,petrovich22temos,chen2023executing,zhang2022motiondiffuse,zhang2023motiongpt,zhang2023generating,tevet2022motionclip,ahuja2019language2pose,guo2022generating,kim2023flame, lu2023humantomato, raab2023single,yonatan2023,dabral2023mofusion,wei2023understanding,zhang2023tedi,kong2023priority,yazdian2023motionscript,barquero2024seamless,zhou2023emdm,ma2024contact,zhong2024smoodi,zhao2024dart,jin2024local,li2024unimotion,dai2024motionlcm,liu2024programmable,yuan2024mogents,zhang2024finemogen} is popular and extended to various applications, including the text-conditioned generation of multiple-person~\cite{liu2023interactive,wang2023intercontrol,ghosh2023remos,li2024two,huang2024closely} and human-scene interaction~\cite{huang2023diffusion,jiang2024scaling, cong2024laserhuman,cen2024generating}. Our goal is to model human and object dynamics concurrently guided by text.

\noindent{\bf Human-Object Interaction Generation.}
Synthesizing hand-object interactions~\cite{li2023task,ye2023affordance,zheng2023cams,zhou2022toch,zhang2023artigrasp,cao2024multi,zhang2024manidext,ma2024madiff,ma2024diff,tian2024gaze,wu2024dice,christen2024diffh2o,liu2023contactgen,cha2024text2hoi} and single-frame human-object interactions~\cite{xie2022chore,zhang2020perceiving,wang2022reconstructing,petrov2023object,hou2023compositional,kim2023ncho,xusemantic,yang2024person,dai2024interfusion} are popular topics and extended to zero-shot settings~\cite{li2024genzi,yang2023lemon,kim2024zero,yang2024egochoir}. 
Recently, researchers explore whole-body dynamic interaction generation, in kinematic-based approaches~\cite{starke2019neural,starke2020local,taheri2022goal,wu2022saga,kulkarni2023nifty,zhang2022couch,lee2023locomotion,xu2021d3dhoi,ghosh2022imos,corona2020context,9714029,razali2023action,Mandery2015a,Mandery2016b,krebs2021kit,xu2023interdiff,li2023object, hassan2021stochastic,wu2024thor,daiya2024collage,wu2024human,song2024hoianimator} and physics-based approaches~\cite{liu2018learning,chao2021learning,merel2020catch,hassan2023synthesizing,bae2023pmp,yang2022learning,xie2022learning,xie2023hierarchical,pan2023synthesizing,braun2023physically,wang2023physhoi, xiao2024unified,cui2024anyskill,tessler2024maskedmimic,liu2024physreaction,wang2024strategy,luo2024smplolympics,tevet2024closd}.
Current methods in HOI synthesis are often restricted by a narrow scope of actions, the use of non-dynamic objects, and a lack of comprehensive whole-body motion.  Our work aims to generate diverse whole-body interactions with various objects, and enables control through language input. Recent datasets~\cite{taheri2020grab,bhatnagar22behave,jiang2022chairs,huang2022intercap,zhang2023neuraldome,fan2023arctic,li2023object,zhao2023im,kim2024parahome,wu2024himo,yang2024f,zhang2024core4d,xie2024intertrack,zhang2024hoi} provide the groundwork for research in this area, and concurrent efforts~\cite{peng2023hoi,diller2023cg,li2023controllable} demonstrate the feasibility of applying supervised learning methods via annotating datasets. However, the amount of data currently available fall short when compared to more extensive text-motion datasets~\cite{mahmood2019amass,guo2022generating,lin2023motionx}. This discrepancy in data volume limits the capability of supervised methods to capture the complexity of human-object interactions, motivating us to investigate the potential of zero-shot generation.

\noindent \textbf{External Knowledge from LLMs.} Large language models (LLMs) are being used for advanced visual tasks, such as editing images based on instructions~\cite{brooks2023instructpix2pix}. In digital humans, they are used to reconstruct 3D human-object interactions~\cite{wang2022reconstructing} and generate human motion~\cite{athanasiou2023sinc,yao2023moconvq,jiang2023motiongpt,zhang2023motiongpt} as well as human-scene interactions~\cite{xiao2023unified}. Our approach is inspired by \cite{wang2022reconstructing}, which uses LLMs to infer contact body parts with a given object for reconstructing 3D human-object interactions -- a task different from ours. Our approach utilizes GPT-4~\cite{chatgpt} or Llama 2~\cite{touvron2023llama}, to not only understand contact body parts but also narrow the distribution gap between different tasks, and provide knowledge for interaction retrieval. This is accomplished by utilizing the in-context learning capabilities of LLMs~\cite{corona2020context} and their support for retrieval-augmented generation~\cite{lewis2020retrieval}.

\section{Methodology}\label{sec:method}
\noindent{\bf Problem Formulation.} Our goal is to synthesize a sequence of 3D human-object interactions \( \boldsymbol{x} \) that satisfies a descriptive text \( p \). This sequence is a series of tuples \( [(\boldsymbol{h}_1, \boldsymbol{o}_1), (\boldsymbol{h}_2, \boldsymbol{o}_2), \ldots, (\boldsymbol{h}_M, \boldsymbol{o}_M)] \), where \( \boldsymbol{h}_i \) represents the human pose parameters defined in the SMPL model~\cite{loper2015smpl}, while the shape of the human is unified the same as~\cite{guo2022generating}. \( \boldsymbol{o}_i \) defines the pose of the rigid object in terms of its 3D spatial position and orientation. The sequence length \( M \) is variable and is dynamically determined by our text-to-motion model based on the input text \( p \). We do \emph{not} require text supervision for training.
\begin{figure*}
    \centering
    \includegraphics[width=\textwidth]{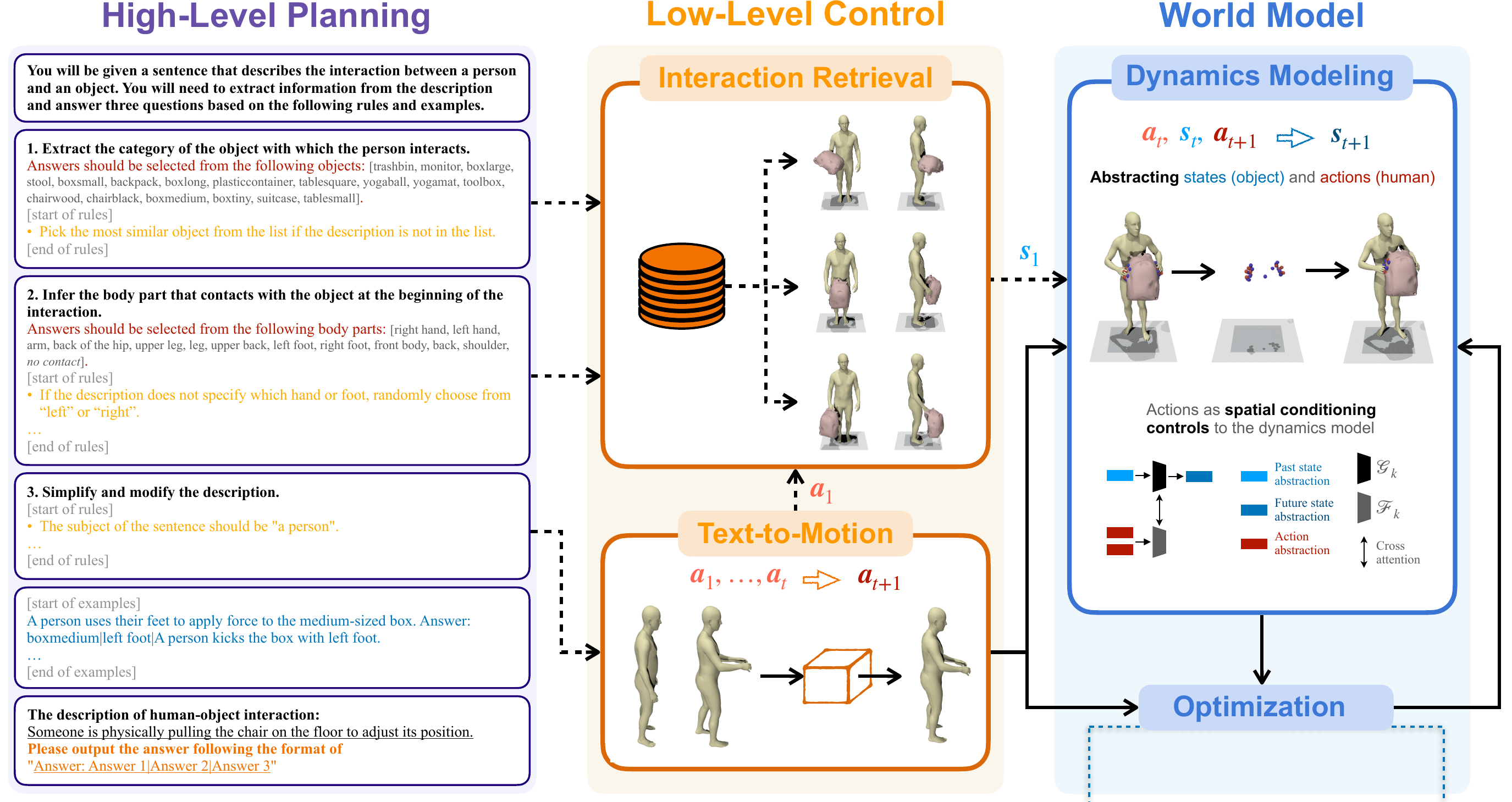}
    \caption{\textbf{An overview of our \ours.}
    \textbf{(i)} Our high-level \high~analyzes the description using LLMs and provides guidance to the low-level \low. \textbf{(ii)} Our low-level \low~includes a text-to-motion model that translates text into human actions $\color{red} \boldsymbol{a}_{t+1}$, and an interaction retrieval model that extracts the object's first state $\color{blue}\boldsymbol{s}_1$. \textbf{(iii)} Our world model executes actions to output the next state $\color{blue}\boldsymbol{s}_{t+1}$. We achieve this by abstracting the problem as predicting the motion of contact vertices -- represented by {\color{red}{red}} spheres for humans and {\color{blue}{blue}} spheres for objects on the top right -- using human vertices as controls for the prediction of object vertices. An optimization process is coupled with the dynamics model, projecting the state and action onto valid counterparts. Solid arrows mean that the process is performed iteratively.}
    \label{fig:method}
\end{figure*}

\noindent{\bf Overview.} 
Our framework, illustrated in Figure \ref{fig:method}, can be conceptualized as a Markov decision process (MDP).
We begin by dividing the motion sequence into \( T \) segments, each with \( m \) frames, where \( M = T \times m \). Object motion \( \{\boldsymbol{o}_i\}_{i=1}^M \) can be seen as a sequence of environmental states \( \{\boldsymbol{s}_t\}_{t=1}^T \), and human motion \( \{\boldsymbol{h}_i\}_{i=1}^M \) is described as a sequence of actions \( \{\boldsymbol{a}_t\}_{t=1}^T \) that interact with the environment. 
Under such an MDP setup, our framework starts with high-level planning \( L \), which deciphers textual interaction description \( p \) by \( g = L(p) \) (Sec.~\ref{sec:high}).  
Then, a text-to-motion model
\( \pi \) translates context \( g \) into human actions, modeled as \( \boldsymbol a_{t+1} \sim \pi(\boldsymbol a_{t+1} | \boldsymbol s_t, \{\boldsymbol a_{i}\}_{i=1}^{t}, g) \) (Sec.~\ref{sec:low}).
The interaction retrieval \( R \) proposes an initial object state \( \boldsymbol s_1 \sim R(\boldsymbol s_{1} | \boldsymbol a_1, g)\), based on the initial action \( \boldsymbol a_1 \) and context \( g \) (Sec.~\ref{sec:low}).
After that, a world model \( P \) is trained to predict future states \( \boldsymbol s_{t+1}  \sim P(\boldsymbol s_{t+1} | \boldsymbol a_t,  \boldsymbol s_t, \boldsymbol a_{t+1})\) from the current action and state (Sec.~\ref{sec:world}).
Our world model incorporates an optimization process,
for both state and action refinement (Sec.~\ref{sec:optim}). 
Notably, the text-to-motion and world models are executed \emph{iteratively} until text-to-motion generates an end frame.

\subsection{High-Level Planning}\label{sec:high}
Leveraging LLMs' strong reasoning capabilities and inherent common sense, our high-level \high~\(L\) yields interaction details \(g = L(p)\) that cannot be na\"ively extracted in textual descriptions $p$. The process undertaken by \(L\) encompasses three steps:
(\textbf{i}) \textit{Determining the object}: the LLM is employed to translate described objects into corresponding categories from a predefined list.
(\textbf{ii}) \textit{Determining initial human-object contact}: the LLM infers the body parts involved in
the interaction, drawing from a list defined in the SMPL model~\cite{loper2015smpl}. And most importantly,
(\textbf{iii}) \textit{reducing the distribution gap}: the LLM bridges the distribution gap between the free-form textual input and the language used within the training data of the text-to-motion model~\cite{guo2022generating}. This involves standardizing syntax and content according to designed guidelines.
In Figure~\ref{fig:method}, we demonstrate the prompt we used with the few-shot prompting~\cite{brown2020language}. We define intermediate thoughts and the final thought, \textit{i.e.}, answers to three questions, as detailed information \(g = L(p)\), which guides the subsequent procedure, structuring the entire framework with a philosophy similar to  retrieval-augmented generation~\cite{lewis2020retrieval}. 
Our high-level \high~operates indirectly in the generation of interactions. Nonetheless, it narrows the vast range of possible interactions in the real world into a more manageable distribution within the capabilities of our framework. We incorporate GPT-4~\cite{chatgpt} and Llama-2~\cite{touvron2023llama} for evaluation.

\subsection{Low-Level Control}\label{sec:low}
With the information \(g\) derived from the description \(p\), the low-level \low~aims to create a sequence of human actions \(\{\boldsymbol{a}_t\}_{t=1}^T\) by a text-to-motion model, and an initial state \(\boldsymbol{s}_1\) by interaction retrieval, such that they correspond to the objectives outlined by \(g\).

\noindent\textbf{Text-to-Motion.} 
We utilize a text-to-motion model \(\pi\) to develop actions to be executed in the world model. At each timestep \( t \), \(\pi\) receives the sequence of previous actions \(\{\boldsymbol{a}_{i}\}_{i=1}^{t}\) and the text tokens encoded from the rewritten description in \( g = L(p) \), and produces a next action \({\boldsymbol{a}_{t+1}}\), which later in Sec.~\ref{sec:optim} will be adjusted through an optimization process that intertwines actions with the object state \(\boldsymbol{s}_t\). Thus, the overall process can be formally defined as \(\boldsymbol{a}_{t+1} \sim \pi(\boldsymbol{a}_{t+1} | \boldsymbol{s}_t, \{\boldsymbol{a}_{i}\}_{i=1}^{t}, g)\), while the initial action \(\boldsymbol{a}_1 \sim \pi(\boldsymbol{a}_1 | g)\) is influenced merely by context \( g \) without prior actions or states, which will be used in interaction retrieval. \(\pi\)  builds upon existing text-to-motion models, where we evaluate MDM~\cite{tevet2022human}, MotionDiffuse~\cite{zhang2022motiondiffuse}, ReMoDiffuse~\cite{zhang2023remodiffuse}, and MotionGPT~\cite{jiang2023motiongpt}.

\noindent\textbf{Interaction Retrieval.} The interaction retrieval component \( R \) establishes the initial state \(\boldsymbol{s}_1 \sim R(\boldsymbol{s}_1 | \boldsymbol{a}_1, g)\), based on the initial action \(\boldsymbol{a}_1\) generated by the text-to-motion model. We propose a user-friendly pipeline for this purpose built on handcrafted rules. First, we create databases by extracting HOI frames from the training sets of each target datasets — BEHAVE~\cite{bhatnagar22behave}, OMOMO~\cite{li2023object}, and CHAIRS~\cite{jiang2022chairs}. The indexing key for retrieval is a tuple consisting of the body part in contact and the category of the involved object. Each retrieval value is a per-frame contact map, represented by a list of \(K\) vertex pairs \(\{(d_h^i, d_o^i)\}_{i=1}^K\). Here, \(d_h^i\) refers to the contact vertex on the human surface, while \(d_o^i\) refers to the corresponding contact vertex on the object surface. This contact map is linked to its corresponding key, creating a searchable record of interactions.
During the inference stage, using the body part and object information provided by the high-level \high~(Sec.~\ref{sec:high}), we retrieve all relevant contact maps from the database. We then sample one map \(\{(d_h^i, d_o^i)\}_{i=1}^K\) and use it to establish the object state \(\boldsymbol{s}_1 \sim R(\boldsymbol{s}_1 | \boldsymbol{a}_1, g)\), thus initializing the interaction. Further details including how we ensure consistency between the sampled state and human action are provided in Sec.~\ref{sec:low_level_supp} of the Appendix. We also discuss an alternative learning-based approach in Sec.~\ref{sec:low_level_supp}.

\subsection{World Model}\label{sec:world}

Our world model combines a dynamics model and the optimization process, dedicated to simulating state transitions affected by applied actions. While drawing inspiration from similar concepts utilized in robotics~\cite{wu2023daydreamer,seo2023masked} and autonomous driving systems~\cite{kim2020learning}, we use it here to generate HOI trajectories. 
This model, trained on the training set of a 3D HOI dataset such as BEHAVE~\cite{bhatnagar22behave}, serves a similar role as a simulator but is much simpler -- it takes the preceding object state \(\boldsymbol s_t\) along with a pair of consecutive actions \(\boldsymbol a_t\) and \(\boldsymbol a_{t+1}\), and predicts the subsequent object state \(\boldsymbol s_{t+1}\). The interplay between the low-level \low~and the world model ultimately produces a coherent interaction rollout.

In designing the dynamics model, a na\"ive method would be directly taking raw actions, states, and object geometry as input. However, this suffers from a severe generalization problem during inference: the dynamics model is likely to encounter human actions and object geometry that do not exist in the training set, since our text-to-motion model is not trained with object interaction data.
To overcome this limitation, we instead focus on encoding interactions through the contact vertices on the object, which capture both the action and object geometry, as shown in Figure~\ref{fig:method}. This \textit{locality} ensures that the dynamics model remains focused on interactions in the contact region, without being distracted by the motion of body parts and geometry details that are irrelevant to the interaction.

\noindent\textbf{Input Representation.} Specifically, at each timestep $t$, we abstract the past actions as $H$ historical vertex trajectories \(\{\{\boldsymbol v_i^j\}_{j=1}^N\}_{i=1}^{H}\), and the future actions as $F=m$ future vertex trajectories \(\{\{\boldsymbol v_i^j\}_{j=1}^N\}_{i=H+1}^{H+F}\), where non-fixed variable $N$ is the number of sampled contact vertices, and $m$ is the length of segments as mentioned in the overview of Sec~\ref{sec:method}. Note that we train our dynamics model to forecast over a longer duration than the past motion (\( F > H \)), only the foremost future action will be used for autoregressive generation during the inference, as suggested in~\cite{chi2023diffusionpolicy}.
To determine these \( N \) vertices, we start with object's signed distance fields \( \{\mathbf{sdf}_i\}_{i=1}^{H} \) over the past \( H \) frames, derived from the past state \( \boldsymbol{s}_t \). We then sample vertices that meet the following criteria:
$|\mathbf{sdf}_i(\boldsymbol v_i^j)| \leq \delta_1, \mathbf{sdf}_i(\boldsymbol v_i^k)| \leq \delta_1, \forall i=1,\ldots,H, \forall j$, and $\|\boldsymbol{v}_i^j - \boldsymbol{v}_i^k\| \geq \delta_2,  \forall j \neq k$,
where \( \delta_1 \) and \( \delta_2 \) are two hyperparameters. The objective is to sparsely sample contact vertices while ensuring that they are sufficient to encompass the interaction.
We characterize each vertex trajectory \(\{\boldsymbol{v}_i^j\}_{i=1}^{H+F}\) with a feature \(\boldsymbol{f}^j\) to provide (i) human vertex coordinates at T-pose, providing information about the position of the human vertex on the body surface; (ii) the vertex-to-object surface vector, indicating vertex's impact on the object as well as inherently including information related to the object's shape; and (iii) the vertex's velocity relative to its nearest object vertex. Thus, the model needs to learn how the features of human action \(\boldsymbol{f}^j\) affect the evolution of the state of the object.

\noindent\textbf{Architecture.} As demonstrated in Figure~\ref{fig:method}, the network comprises two components: $\mathcal{G}$ that operates without contact vertex conditions, applicable in scenarios where no contact occurs, and $\mathcal{F}$, which incorporates contact vertex conditions into the object trajectory when contact is present. The k-th layer of $\mathcal{G}$ can be initiated as \(\mathcal G_k(\boldsymbol x_{k}, \boldsymbol \Theta)\), mapping the input feature map \(\boldsymbol x_k\) at the $k$-th layer to another feature map, with \(\Theta\) denoting the MLP's parameters. To incorporate human vertex controls, we introduce a second network \(\mathcal{F}_k(\boldsymbol{y}^j_k, \boldsymbol \Theta_v)\) operating on $N$ vertex features $\{\boldsymbol{y}^j_k\}_{j=1}^N$, where $\boldsymbol \Theta_v$ is its parameters. With a cross-attention layer $\mathrm{Attn}$, a dynamics block is formulated as: $\boldsymbol x_{k+1}, \{\boldsymbol{y}^j_{k+1}\}_{j=1}^N = \mathrm{Attn} (\mathcal G_k(\boldsymbol x_k, \boldsymbol \Theta), \{\mathcal{F}_k(\boldsymbol{y}^j_k, \boldsymbol \Theta_v)\}_{j=1}^N).$
We stack multiple dynamics blocks to form the model. The initial input, \(\boldsymbol x_0\), corresponds to the previous state \(\boldsymbol s_t\), while each \(\boldsymbol{y}^j_0\) represents the feature of the vertex trajectory, containing both the trajectory \(\{\boldsymbol{v}_i^j\}_{i=1}^{H+F}\) and its associated feature vector \(\boldsymbol{f}^j\). The output of this model is preliminary and subject to further optimization as introduced in Sec.~\ref{sec:optim}, which will yield the final future state. We utilize the Mean Squared Error loss to train the dynamics model. For more explanations, please refer to Sec.~\ref{sec:world_supp} of the Appendix.

\subsection{Optimization}\label{sec:optim}
Optimization plays a role in introducing prior knowledge and avoiding the accumulation of errors. During inference, we input the action $\boldsymbol{a}_{t+1}$ and state $\boldsymbol{s}_{t+1}$ and refine them. This refinement is achieved through gradient descent on the human and object pose parameters. Our optimization includes several loss terms: a fitting loss to align optimized results with their preliminary one, a velocity loss for temporal smoothness, a contact loss to promote occurring contact, and a collision loss to reduce penetration. We provide detailed formulations in Sec.~\ref{sec:optim_supp} of the Appendix.

\begin{table}[t]
    \centering
    \caption{\textbf{Quantitative results} on evaluating the dynamics model. Our dynamics model with vertex-based action generates interactions of the best quality.}
    \label{tab:t2m_after}
    \resizebox{\textwidth}{!}{
    \begin{tabular}{@{}lccccccc@{}}
        \toprule
        \multirow{2}{*}{Methods} & &\multicolumn{2}{c}{Text-to-Interaction} & & \multicolumn{3}{c}{Interaction Prediction~\cite{xu2023interdiff}} \\
        
        \cmidrule(lr){2-4} \cmidrule(lr){5-8} 
        & &  CMD $\downarrow$  & Pene. ($10^{-2}\%$) $\downarrow$ & & Trans. Err. (mm) $\downarrow$  & Rot. Err. ($10^{-3}$ rad) $\downarrow$  & Pene. ($10^{-2}\%$) $\downarrow$ \\
        
        \midrule
        w/o action                   &   &    0.424   &   533 & &  123         &    256       &  228  \\
        contact markers as action (InterDiff~\cite{xu2023interdiff})           &                      &  0.219     &  484   & &  123        &    226      &  164  \\
        human motion as action  &   &            0.325  &  957  & &   129       &    265   &   218   \\ 
        contact vertices as action (\textbf{ours})  &       &          \textbf{0.151}   &   \textbf{443}  & &  \textbf{119}        &   \textbf{221}    &    \textbf{156}    \\
        
        \bottomrule
    \end{tabular}}
    \vspace{-1.2em}
\end{table}
\begin{table}
    \centering
    \caption{\textbf{Quantitative results} on human motion quality given our annotation on the BEHAVE~\cite{bhatnagar22behave} dataset. We show that our high-level planning effectively adapts single human generators into human-object interaction generation. To evaluate R-Precision, a batch size of 16 is selected.}
    \label{tab:t2m_planning}
    \resizebox{\textwidth}{!}{
    \begin{tabular}{@{}lcccccccc@{}}
        \toprule
        \multirow{2}{*}{Methods} & \multirow{2}{*}{\makecell{Planning \\ (\textbf{Ours})}} & \multicolumn{3}{c}{R-Precision$^\uparrow$} & \multirow{2}{*}{FID$^\downarrow$} & \multirow{2}{*}{MM Dist$^\downarrow$} & \multirow{2}{*}{Multimodality$^\uparrow$} & \multirow{2}{*}{Diversity$^\rightarrow$} \\
        
        \cmidrule(lr){3-5} 
        & & Top 1 & Top 2 & Top 3 & & &  \\
        \midrule
        Ground Truth & - & 0.237$^{\pm0.004}$ & 0.392$^{\pm0.004}$ & 0.496$^{\pm0.005}$ & 0.024$^{\pm0.000}$ & 4.259$^{\pm0.006}$& - & 6.510$^{\pm0.227}$  \\
        \midrule
        \multirow{2}{*}{MDM~\cite{tevet2022human}}& $\times$ & 0.153$^{\pm0.016}$ & 0.279$^{\pm0.026}$ & 0.398$^{\pm0.016}$ & 12.279$^{\pm0.217}$ & 5.351$^{\pm0.057}$&\textbf{7.604}$^{\pm0.190}$ & 7.598$^{\pm0.334}$  \\
        & $\checkmark$ & \textbf{0.163}$^{\pm0.010}$ & \textbf{0.307}$^{\pm0.043}$ & \textbf{0.402}$^{\pm0.019}$ & 
        \textbf{10.374}$^{\pm0.304}$ & \textbf{5.303}$^{\pm0.117}$& 7.281$^{\pm0.083}$ & \textbf{7.471}$^{\pm0.427}$  \\
        \midrule 

        \multirow{2}{*}{MotionDiffuse~\cite{zhang2022motiondiffuse}} & $\times$ & 0.205$^{\pm0.011}$ & 0.351$^{\pm0.002}$ & 0.458$^{\pm0.021}$ & 10.208$^{\pm0.500}$ & 4.837$^{\pm0.064}$ & 4.520$^{\pm0.163}$ & 7.323$^{\pm0.412}$  \\
        & $\checkmark$ & \textbf{0.216}$^{\pm0.032}$ & \textbf{0.369}$^{\pm0.023}$ & \textbf{0.472}$^{\pm0.027}$ & \textbf{9.015}$^{\pm0.403}$ & \textbf{4.649}$^{\pm0.029}$& \textbf{4.991}$^{\pm0.172}$ & \textbf{7.295}$^{\pm0.501}$  \\
        \midrule
        
        \multirow{2}{*}{ReMoDiffuse~\cite{zhang2023remodiffuse}} & $\times$ & 0.196$^{\pm0.009}$ & 0.338$^{\pm0.011}$ & 0.448$^{\pm0.012}$ & 6.385$^{\pm0.201}$ & 4.855$^{\pm0.029}$ &5.889$^{\pm0.524}$ & \textbf{7.160}$^{\pm0.306}$  \\

         & $\checkmark$ & \textbf{0.223}$^{\pm0.006}$ & \textbf{0.368}$^{\pm0.015}$ & \textbf{0.482}$^{\pm0.011}$ & \textbf{5.237}$^{\pm0.174}$ & \textbf{4.784}$^{\pm0.053}$&\textbf{6.350}$^{\pm0.411}$ & 7.201$^{\pm0.318}$  \\
        \midrule
        
        \multirow{2}{*}{MotionGPT~\cite{jiang2023motiongpt}} & $\times$ & 0.233$^{\pm0.003}$ & 0.344$^{\pm0.004}$ & 0.457$^{\pm0.005}$ & 5.497$^{\pm0.106}$ & 5.205$^{\pm0.027}$&1.062$^{\pm0.211}$ & 8.316$^{\pm0.204}$  \\
        
         & $\checkmark$ & \textbf{0.234}$^{\pm0.004}$ & \textbf{0.387}$^{\pm0.003}$ & \textbf{0.471}$^{\pm0.007}$ & \textbf{4.751}$^{\pm0.121}$ & \textbf{4.995}$^{\pm0.003}$ &\textbf{1.337}$^{\pm0.193}$  & \textbf{7.106}$^{\pm0.487}$  \\

        \bottomrule
    \end{tabular}}
    \vspace{-1.2em}
\end{table}
\begin{table}
    \centering
    \caption{\small\textbf{Quantitative results} on human motion quality on the OMOMO~\cite{li2023object} dataset with their provided annotation. We show that our high-level planning narrows the distribution gap and adapts single human generators into human-object interaction generation. To evaluate R-Precision, a batch size of 32 is selected.}
    \label{tab:omomo_planning}
    \resizebox{\textwidth}{!}{
    \begin{tabular}{@{}lcccccccc@{}}
        \toprule
        \multirow{2}{*}{Methods} & \multirow{2}{*}{\makecell{Planning \\ (\textbf{Ours})}} & \multicolumn{3}{c}{R-Precision$^\uparrow$} & \multirow{2}{*}{FID$^\downarrow$} & \multirow{2}{*}{MM Dist$^\downarrow$} & \multirow{2}{*}{Multimodality$^\uparrow$} & \multirow{2}{*}{Diversity$^\rightarrow$} \\
        
        \cmidrule(lr){3-5} 
        & & Top 1 & Top 2 & Top 3 & & &  \\
        \midrule
        Ground Truth & - & 0.044$^{\pm0.004}$ & 0.095$^{\pm0.008}$ & 0.151$^{\pm0.009}$ & 0.000$^{\pm0.000}$ & 6.858$^{\pm0.006}$& - & 5.660$^{\pm0.110}$  \\
        \midrule
        \multirow{2}{*}{MDM~\cite{tevet2022human}}& $\times$ & 0.056$^{\pm0.005}$ & 0.096$^{\pm0.007}$ & 0.135$^{\pm0.006}$ & 16.638$^{\pm0.631}$ & 7.110$^{\pm0.063}$&2.446$^{\pm0.456}$ & \textbf{5.862}$^{\pm0.520}$  \\
        & $\checkmark$ & \textbf{0.062}$^{\pm0.006}$ & \textbf{0.109}$^{\pm0.004}$ & \textbf{0.155}$^{\pm0.008}$ & 
        \textbf{15.735}$^{\pm0.285}$ & \textbf{6.889}$^{\pm0.082}$&\textbf{2.663}$^{\pm0.520}$ & 6.461$^{\pm0.841}$  \\
        \midrule 

        \multirow{2}{*}{MotionDiffuse~\cite{zhang2022motiondiffuse}} & $\times$ & 0.048$^{\pm0.006}$ & 0.094$^{\pm0.008}$ & 0.143$^{\pm0.013}$ & 15.442$^{\pm0.231}$ & \textbf{5.799}$^{\pm0.054}$ & 1.658$^{\pm0.209}$ & 5.981$^{\pm0.516}$  \\
        & $\checkmark$ & \textbf{0.075}$^{\pm0.005}$ & \textbf{0.141}$^{\pm0.015}$ & \textbf{0.189}$^{\pm0.009}$ & \textbf{10.815}$^{\pm0.093}$ & 5.916$^{\pm0.094}$& \textbf{1.677}$^{\pm0.264}$ & \textbf{5.718}$^{\pm0.522}$  \\
        \midrule
        
        \multirow{2}{*}{ReMoDiffuse~\cite{zhang2023remodiffuse}} & $\times$ & 0.062$^{\pm0.003}$ & 0.111$^{\pm0.005}$ & 0.160$^{\pm0.012}$ & 15.479$^{\pm0.209}$ & 5.690$^{\pm0.049}$ &1.179$^{\pm0.145}$ & 6.032$^{\pm0.540}$  \\

         & $\checkmark$ & \textbf{0.067}$^{\pm0.004}$ & \textbf{0.127}$^{\pm0.006}$ & \textbf{0.174}$^{\pm0.006}$ & \textbf{14.560}$^{\pm0.080}$ & \textbf{5.678}$^{\pm0.033}$&\textbf{1.193}$^{\pm0.202}$ & \textbf{5.368}$^{\pm0.417}$  \\
        \midrule
        
        \multirow{2}{*}{MotionGPT~\cite{jiang2023motiongpt}} & $\times$ & 0.061$^{\pm0.005}$ & 0.114$^{\pm0.006}$ & 0.152$^{\pm0.006}$ & 18.472$^{\pm0.528}$ & 6.358$^{\pm0.076}$&\textbf{4.553}$^{\pm0.244}$ & \textbf{6.726}$^{\pm0.156}$  \\
        
         & $\checkmark$ & \textbf{0.064}$^{\pm0.007}$ & \textbf{0.121}$^{\pm0.007}$ & \textbf{0.164}$^{\pm0.009}$ & \textbf{17.512}$^{\pm0.498}$ & \textbf{6.287}$^{\pm0.041}$ &4.470$^{\pm0.191}$  & 7.048$^{\pm0.169}$  \\

        \bottomrule
    \end{tabular}}

\end{table}
\begin{figure*}
    \centering
    \includegraphics[width=\textwidth]{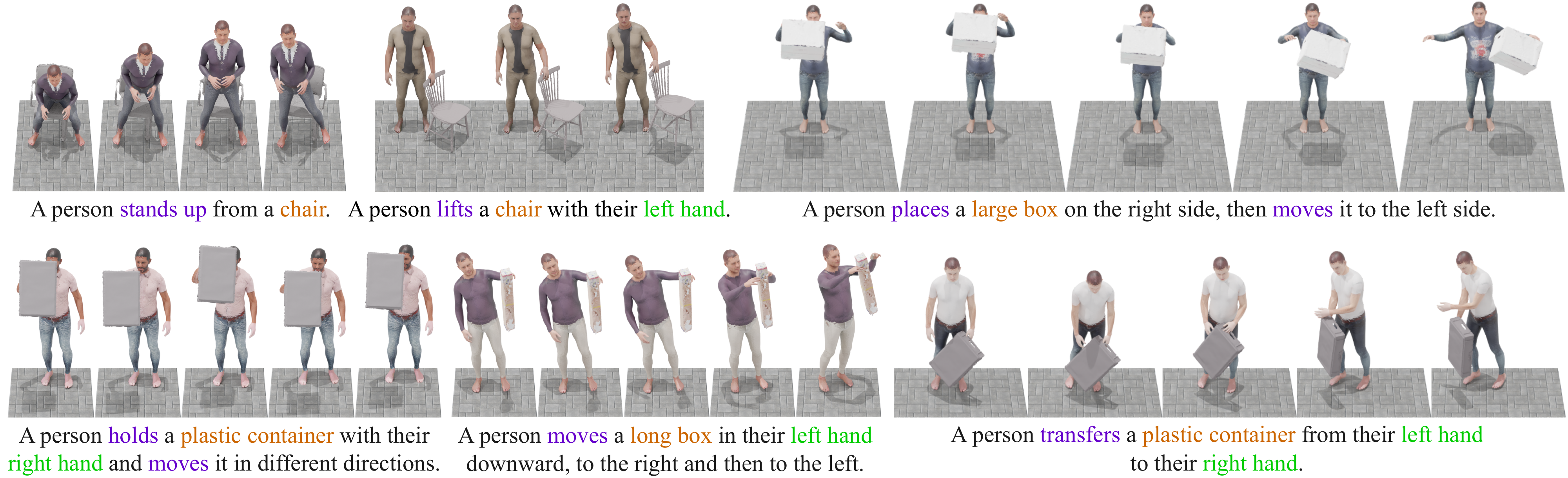}
    \caption{\textbf{Qualitative results} on free-form text input. The interaction sequences, with textures from~\cite{casas2023smplitex}, are presented through a time-series visualization. 
    }
    \label{fig:qual1}
\end{figure*}

\section{Experiments}\label{sec:exp}
Extensive comparisons evaluate the performance of our \ours~across two motion-relevant tasks. Details of the evaluation settings are provided in Sec.~\ref{sec:exp_set}. We present both quantitative (Sec.~\ref{sec:quan}) and qualitative (Sec.~\ref{sec:qual}) results for our approach. Additionally, we perform ablation studies to verify the efficacy of each component within our framework. These studies also cover the interaction prediction task~\cite{xu2023interdiff} to evaluate our dynamics model. Additional details and results are presented in Sec.~\ref{sec:imple_supp} and Sec.~\ref{sec:qual_supp} of the Appendix. Please refer to our \href{https://sirui-xu.github.io/InterDreamer/}{website} for video results.
\begin{figure*}
    \centering
    \includegraphics[width=\textwidth]{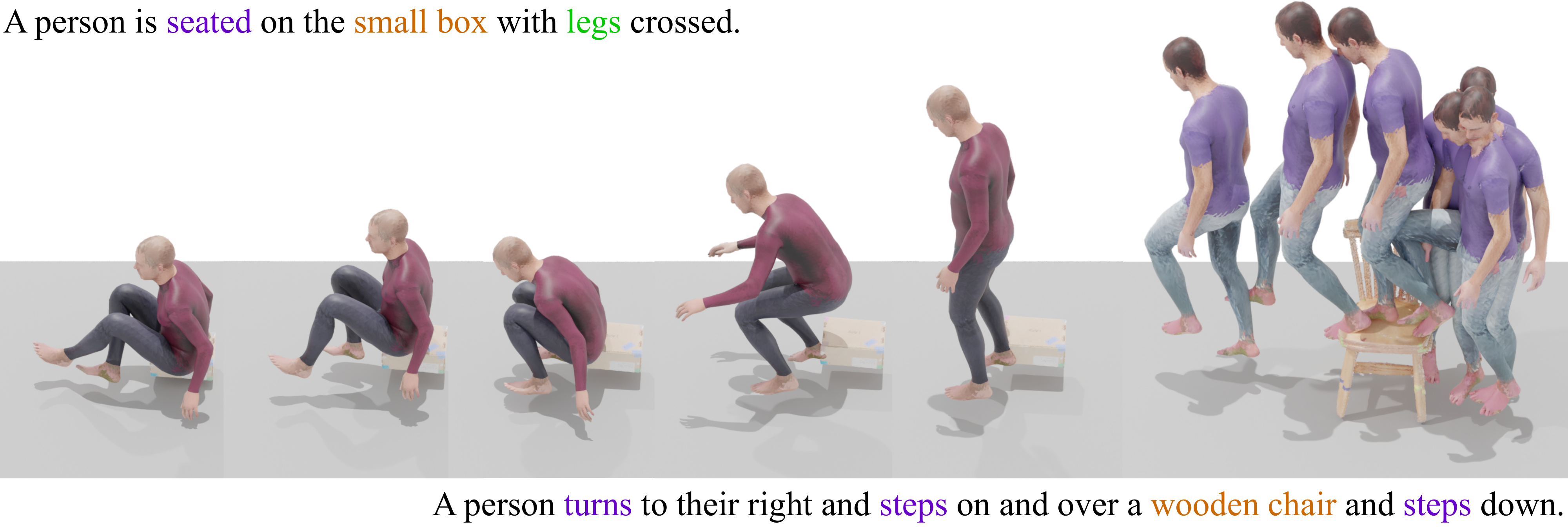}
    \caption{\textbf{Qualitative results} in more challenge scenarios with \textit{free-form input not} from our annotations, showing the ability of our \ours~to fit \textit{object sizes} and handle \textit{complex and long sequences}. Here, our synergized models are GPT-4~\cite{chatgpt} and MotionGPT~\cite{jiang2023motiongpt}.
    }
    \label{fig:qual_ood}
\end{figure*}
\subsection{Experimental Setup}\label{sec:exp_set}
\noindent \textbf{Datasets.}
We use BEHAVE~\cite{bhatnagar22behave}, CHAIRS~\cite{jiang2022chairs}, and OMOMO~\cite{li2023object} datasets for quantitative evaluation. The BEHAVE dataset includes recordings of 8 individuals interacting with 20 everyday objects, and our analysis focuses on 18 objects for which interaction sequences are available at 30 Hz.
The human pose is modeled using SMPL-H~\cite{MANO}, with hand poses set to an average pose \emph{due to the absence of detailed hand pose in the dataset}. We manually segment the long interaction sequences in the test set, and annotate them with descriptions as well as their starting and ending indices, leading to $532$ subsequences for evaluation. 
For qualitative evaluation, we go beyond using annotations specifically created and employ free -- form text to demonstrate our model's capability on out-of-distribution inputs. To assess our model's performance with novel objects, we expand our retrieval database to include objects from the OMOMO~\cite{li2023object} and CHAIRS~\cite{jiang2022chairs} datasets, while we do not fine-tune the dynamics model on them--a direct qualitative evaluation without additional adaptation.
\begin{figure*}
    \centering
    \includegraphics[width=\textwidth]{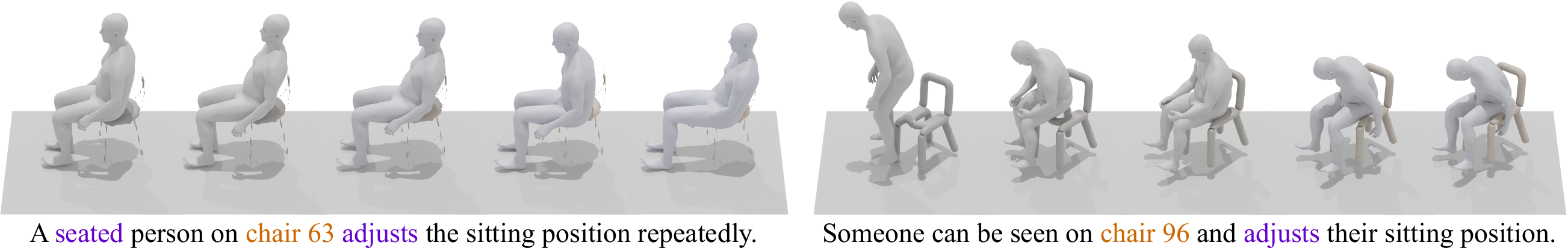}
     \caption{\textbf{Qualitative results} on the CHAIRS~\cite{jiang2022chairs} dataset. Our dynamics model trained on the BEHAVE~\cite{bhatnagar22behave} dataset generalizes well on the CHAIRS objects unseen in training. Frames are separately visualized. Here, our synergized models are GPT-4~\cite{chatgpt} and MotionGPT~\cite{jiang2023motiongpt}.} 
    \label{fig:chairs}
\end{figure*}
\begin{figure}
    \centering
    \includegraphics[width=\columnwidth]{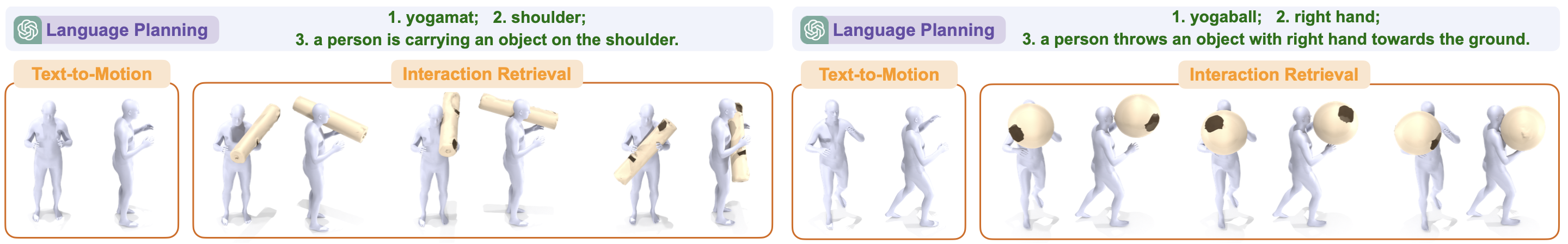}
    \caption{Results from the interaction retrieval. We demonstrate that our proposed retrieval approach based on handcraft rules can extract diverse and realistic interactions. %Of these, one interaction is sampled for subsequent steps.
    }
    \label{fig:retrieval}
\end{figure}
\begin{figure}
    \centering
    \includegraphics[width=\textwidth]{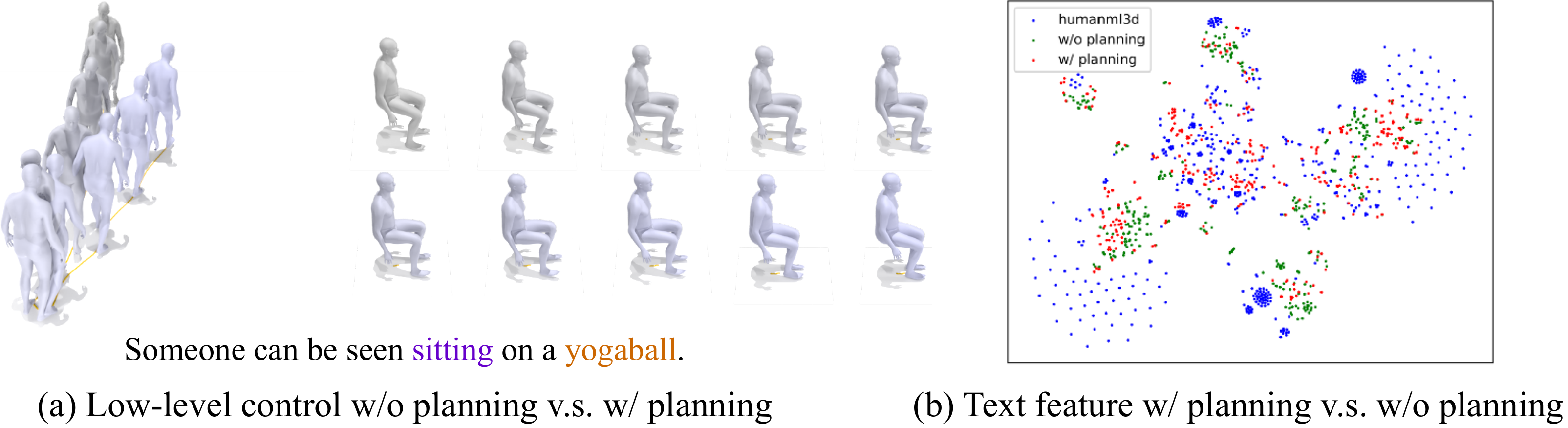}
    \caption{(\textbf{a}) \textbf{Ablation study} on the high-level \high. On the \textit{left} are results from MotionGPT~\cite{jiang2023motiongpt} using free-form descriptions, and on the \textit{right} are results with our planning additionally. Without planning, the motion generative model struggles to interpret free-form HOI descriptions and generate semantically-aligned motion. (\textbf{b}) We visualize CLIP~\cite{radford2021learning} features of text on HumanML3D~\cite{guo2022generating} via t-SNE~\cite{van2008visualizing}, raw HOI descriptions (``w/o planning''), and HOI descriptions processed through our high-level planning (``w/ planning''). See Table~\ref{table:hpq} for quantitative measurements.
    }
    \label{fig:motiongpt}
\end{figure}

\noindent \textbf{Metrics.}
The evaluation metrics are divided into three categories:
(\textbf{i}) \textit{Human motion quality}: The Fr\'echet Inception Distance (\textbf{FID}) measures the distance between the generated motion and ground truth. The MultiModality (\textbf{Multimodality}) and \textbf{Diversity} metrics assess the variance in generated human motion. \textbf{R-Precision} evaluates the consistency between the text and the generated human motion within the latent space. MultiModal distance (\textbf{MM Dist}) is the distance between the motion feature and the text feature. We follow~\cite{guo2022generating} to generate motion and text features.
(\textbf{ii}) \textit{Interaction quality}: We propose \textbf{CMD} to measure the distance between contact maps of real interactions and those generated. The per-sequence contact map is defined by the percentage of time that each body part is actively in contact. The detailed formulation is provided in Sec.~\ref{sec:imple_supp} of the Appendix. We also measure the collision (\textbf{Pene.}~\cite{xu2023interdiff}), which calculates the average percentage of object vertices that have non-negative values in the human signed distance fields~\cite{park2019deepsdf}.
(\textbf{iii}) \textit{Object motion accuracy}: The dynamics model's performance in the interaction prediction task~\cite{xu2023interdiff} is evaluated by the accuracy of predicted object motion, including \textbf{Trans.~Err.}, the average distance between predicted and ground truth, and \textbf{Rot.~Err.}, the average distance between the predicted and ground truth. 

\noindent \textbf{Baselines.} Most recent research on text-to-HOI synthesis follows a supervised learning approach~\cite{peng2023hoi, diller2023cg}, making direct quantitative comparisons unfair. Therefore, we primarily focus on qualitative comparisons with these methods. To enable quantitative evaluation, we develop a range of baselines to assess both the overall performance of our framework and the effectiveness of its individual components. In the context of high-level planning, we utilize GPT-4~\cite{chatgpt} and Llama-2~\cite{touvron2023llama}, illustrating the effectiveness of our prompts across different language models. For low-level motion generation control, our baselines include MDM~\cite{tevet2022human}, MotionDiffuse~\cite{zhang2022motiondiffuse}, ReMoDiffuse~\cite{zhang2023remodiffuse}, and MotionGPT~\cite{jiang2023motiongpt}, which span a range of text-to-motion approaches trained on HumanML3D~\cite{guo2022generating} and show the generalizability of our framework. 
To evaluate the dynamics model, we include different design choices: (\textbf{i}) unconditional dynamics model which operates object dynamics independently of human motion; (\textbf{ii}) using human marker features as actions to the dynamics model, similar to~\cite{xu2023interdiff}; (\textbf{iii}) using unprocessed human motion and object pointcloud motion as input to the dynamics model; (\textbf{iv}) our proposed vertex-based actions where only the contacting vertices are used for control. 

\subsection{Quantitative Results}\label{sec:quan}
In Table~\ref{tab:t2m_after}, comparing our framework to baselines with unconditional dynamics model, \ours~achieves better interaction quality in terms of CMD and penetration scores, showing the importance of human influence on object motion. 
Against methods that utilize direct raw human motion or markers for action features, our method demonstrates enhanced performance by offering more fine-grained guidance and extracting generalizable features for dynamics modeling.
Tables~\ref{tab:t2m_planning} and~\ref{tab:omomo_planning} present a comparative analysis of our approach of combining high-level planning with low-level control, where we adopt various text-to-motion models against their counterparts without high-level planning on the BEHAVE and OMOMO datasets. Our approach consistently outperforms baselines. Specifically, \ours~exhibits superior motion quality, reflected by a lower FID, higher R-Precision, and better diversity, highlighting the benefits of incorporating our planning to reduce the distribution gap for the motion generator to generalize in the HOI synthesis task. 
\begin{figure}
    \centering
    \includegraphics[width=\columnwidth]{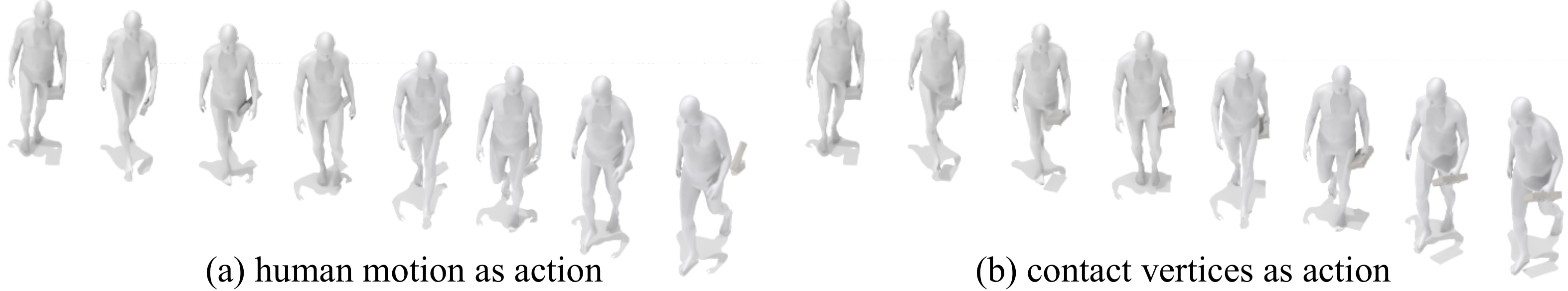}
    \caption{\textbf{Ablation study} on the dynamics model. Given the text description of ``A person walks clockwise while holding a small box with left hand,'' our (\textbf{b}) vertex-based control  can synthesize consistent contacts, which (\textbf{a}) the baseline fails to do.
    }
    \label{fig:control}
\end{figure}

\subsection{Qualitative Results}\label{sec:qual}
Figure~\ref{fig:qual1} displays several results guided by the free-form text. Our method exhibits proficiency in interpreting the textual input and synthesizing dynamic, realistic interactions, despite the absence of training with text-interaction paired data. More importantly, as illustrated in Figure~\ref{fig:qual_ood}, we selectively use more complex sequences of interactive descriptions that are \textit{beyond the scope of the existing HOI dataset}. Figure~\ref{fig:chairs} further exemplifies our method that is able to generalize effectively to the CHAIRS dataset, despite our dynamics model not being trained on it. Figure~\ref{fig:retrieval} depicts the retrieval procedure, resulting in a diverse set of interactions that are both high-quality and semantically aligned. More experimental results and the user study are presented in Sec.~\ref{sec:qual_supp} of the Appendix.

\begin{minipage}[t]{0.6\textwidth}
    \centering
    \captionof{table}{\textbf{Ablation study} on the high-level planning. Q1 and Q2 ask to identify the object category and the contact body part, respectively. We assess the accuracy by comparing the LLM's responses with labels we annotate. Note that the text input to LLMs may contain ambiguities; for example, the annotation is ``hand'' when the motion uses ``right hand.'' We include Q1 Acc$^\ast$ and Q2 Acc$^\ast$ excluding ambiguous text.}
    \label{table:llm}
    \resizebox{\textwidth}{!}{
    \begin{tabular}{ccccccccc}
    \hline\hline
    {LLM ($\#$ of parameters)} & &  Q1 Acc $\uparrow$  & Q1 Acc$^\ast$ $\uparrow$  & Q2 Acc $\uparrow$ & Q2 Acc$^\ast$ $\uparrow$  \\ \hline
    GPT-4~\cite{chatgpt}                  & &   $0.801$        &    $0.997$     &    $0.703$     &   $0.964$   \\
    Llama-2 (7B)~\cite{touvron2023llama}      & &  $0.073$  &     $0.147$       &      $0.436$       &  $0.689$   \\
    Llama-2 (13B)~\cite{touvron2023llama}      & &   $0.232$ &      $0.319$       &       $0.662$    &   $0.853$  \\
    Llama-2 (70B)~\cite{touvron2023llama}       & & $0.722$  & $0.967$                     &   $0.798$ & $0.907$  \\
    \hline\hline
    \end{tabular}}
\end{minipage}
\hfill
\begin{minipage}[t]{0.38\textwidth}
    \centering
    \captionof{table}{\textbf{Quantitative comparison} of text similarity. The text processed by high-level planning is more similar to text in HumanML3D~\cite{guo2022generating} on average, while addressing the distributional gap significantly for challenging out-of-distributional descriptions, compared to text without planning.}
    \label{table:hpq}
    \resizebox{\textwidth}{!}{
    \begin{tabular}{ccccccc}
    \hline\hline
    Sim. to~\cite{guo2022generating}$\uparrow$ & & Average & Out-of-Dist. \\ \hline
    w/o planning                & &   $0.913$        &   $0.838$      \\
    w/ planning      & &  $\mathbf{0.932}$  &     $\mathbf{0.927}$         \\
    \hline\hline
    \end{tabular}}
    \vspace{-2em}
\end{minipage}
\subsection{Ablation Study}\label{sec:ablation}
\noindent \textbf{Adaptability of High-Level Planning.} Is our framework adaptable across different large language models (LLMs)? As illustrated in Table~\ref{table:llm}, our analysis contains two types of language models: GPT-4~\cite{chatgpt}, which is accessible through APIs and operates as a black box model; and Llama-2~\cite{touvron2023llama}, an open-source model. We demonstrate that language models with large parameters exhibit very high accuracy in responding to questions tailored to our prompts, validating the framework's adaptability.

\noindent \textbf{Effectiveness of High-Level Planning with Low-Level Control.}
In consistency with Table~\ref{tab:t2m_planning}, Figure~\ref{fig:motiongpt} offers a qualitative comparison of text-to-motion results, contrasting results with and without LLM-revised text descriptions. The comparison shows that motion generated without LLM-enhanced descriptions often fails to correspond to the intended text, if the text is too challenging, \textit{e.g.,} not in the distribution of HumanML3D~\cite{guo2022generating}, which is used to train text-to-motion models. This underscores the LLM's critical role in bridging the distribution gap. In Figure~\ref{fig:motiongpt}(b), we visualize the CLIP~\cite{radford2021learning} features of descriptions from HumanML3D, our raw annotations, and those processed by high-level planning. Quantitative evidence is provided in Table~\ref{table:hpq}. Text processed through high-level planning demonstrates greater similarity to the HumanML3D dataset. Additionally, we test on more challenging out-of-distribution text, selecting examples with an average cosine similarity to HumanML3D text of less than $0.85$. High-level planning successfully rephrases these texts, significantly increasing their similarity. For example, in Figure~\ref{fig:motiongpt}(a), the text ``Someone can be seen sitting on a yoga ball'' has a cosine similarity of $0.874$ to the closest in-distribution text, while the rephrased text after planning, ``A person is seated on an object,'' achieves a similarity of $0.958$.

\noindent \textbf{Effectiveness of World Model.}
In the quantitative evaluation, we show that the performance of our framework is enhanced by the tailored design of our world model. Table~\ref{tab:t2m_after} provides additional evidence of the effectiveness by integrating the proposed world model, as interaction correction within the InterDiff framework~\cite{xu2023interdiff} in the interaction prediction task. This implementation demonstrates enhanced conditionality in the object dynamics modeling across two tasks, attributed to the vertex-level condition as actions. Doing so effectively removes the whole-body complexity, most of which tends to be irrelevant to the interaction. Figure~\ref{fig:control} further indicates that our vertex-based condition can establish consistent interactions over time, while the condition by raw motion is not robust.
\section{Conclusion} \label{sec:conclusion}
We tackle the task of text-guided 3D human-object interaction generation, aiming to accomplish this without relying on paired text-interaction data.
To this end, we present \ours~that decouples interaction dynamics from semantics, formulating the task as retrieval-augmented generation and Markov decision process, where high-level planning and low-level control are introduced to generate semantically aligned human motion and initial object pose, while a world model is responsible for the object dynamics guided by the interaction. Our approach demonstrates effectiveness in this novel task, suggesting its potential for various real-world applications. 

\noindent \textbf{Limitations.} The current utilization of dynamics modeling could be enhanced. A prospective improvement involves incorporating model-based learning techniques, which empower the agent to more effectively interact and learn a broader range of skills. The results may not be physically plausible and lead to artifacts in some cases, for example, foot skating. Hand poses are rough because they are missing from the dataset, but could be improved by integrating a physics simulator.

\newpage
\section*{Acknowledgments}
We thank Qiusi Zhan for supporting the implementation and evaluation of high-level planning, Jiangwei Yu and Xiyan Xu for their efforts in dataset annotation, and Xiang Li for the discussions. This work was supported in part by NSF Grant 2106825, NIFA Award 2020-67021-32799, the IBM-Illinois Discovery Accelerator Institute, the Toyota Research Institute, and the Jump ARCHES endowment through the Health Care Engineering Systems Center at Illinois and the OSF Foundation. This work used computational resources on NCSA Delta and PTI Jetstream2 through allocations CIS220014, CIS230012, CIS230013, and CIS240311 from the Advanced Cyberinfrastructure Coordination Ecosystem: Services \& Support (ACCESS) program, and on TACC Frontera through the National Artificial Intelligence Research Resource (NAIRR) Pilot.

\bibliographystyle{splncs04}
\bibliography{main}

@inproceedings{dai2024interfusion,
  title={{InterFusion}: Text-Driven Generation of 3D Human-Object Interaction},
  author={Dai, Sisi and Li, Wenhao and Sun, Haowen and Huang, Haibin and Ma, Chongyang and Huang, Hui and Xu, Kai and Hu, Ruizhen},
  booktitle={ECCV},
  year={2024}
}

@inproceedings{cha2024text2hoi,
  title={{Text2HOI}: Text-guided 3D Motion Generation for Hand-Object Interaction},
  author={Cha, Junuk and Kim, Jihyeon and Yoon, Jae Shin and Baek, Seungryul},
  booktitle={CVPR},
  year={2024}
}

@inproceedings{liu2023contactgen,
  title={Contactgen: Generative contact modeling for grasp generation},
  author={Liu, Shaowei and Zhou, Yang and Yang, Jimei and Gupta, Saurabh and Wang, Shenlong},
  booktitle={ICCV},
  year={2023}
}

@inproceedings{hassan2021stochastic,
  title={Stochastic scene-aware motion prediction},
  author={Hassan, Mohamed and Ceylan, Duygu and Villegas, Ruben and Saito, Jun and Yang, Jimei and Zhou, Yi and Black, Michael J},
  booktitle={ICCV},
  year={2021}
}

@inproceedings{lewis2020retrieval,
  title={Retrieval-augmented generation for knowledge-intensive nlp tasks},
  author={Lewis, Patrick and Perez, Ethan and Piktus, Aleksandra and Petroni, Fabio and Karpukhin, Vladimir and Goyal, Naman and K{\"u}ttler, Heinrich and Lewis, Mike and Yih, Wen-tau and Rockt{\"a}schel, Tim and others},
  booktitle={NeurIPS},
  year={2020}
}

@inproceedings{yuan2024mogents,
  title={MoGenTS: Motion Generation based on Spatial-Temporal Joint Modeling},
  author={Yuan, Weihao and Shen, Weichao and He, Yisheng and Dong, Yuan and Gu, Xiaodong and Dong, Zilong and Bo, Liefeng and Huang, Qixing},
  booktitle={ECCV},
  year={2024}
}

@inproceedings{
xiao2024unified,
title={Unified Human-Scene Interaction via Prompted Chain-of-Contacts},
author={Zeqi Xiao and Tai Wang and Jingbo Wang and Jinkun Cao and Wenwei Zhang and Bo Dai and Dahua Lin and Jiangmiao Pang},
booktitle={ICLR},
year={2024},
}

@inproceedings{zhang2024finemogen,
  title={Finemogen: Fine-grained spatio-temporal motion generation and editing},
  author={Zhang, Mingyuan and Li, Huirong and Cai, Zhongang and Ren, Jiawei and Yang, Lei and Liu, Ziwei},
  booktitle={NeurIPS},
  year={2024}
}

@article{tevet2024closd,
  title={{CLoSD}: Closing the Loop between Simulation and Diffusion for multi-task character control},
  author={Tevet, Guy and Raab, Sigal and Cohan, Setareh and Reda, Daniele and Luo, Zhengyi and Peng, Xue Bin and Bermano, Amit H and van de Panne, Michiel},
  journal={arXiv preprint arXiv:2410.03441},
  year={2024}
}

@article{luo2024smplolympics,
  title={{SMPLOlympics}: Sports Environments for Physically Simulated Humanoids},
  author={Luo, Zhengyi and Wang, Jiashun and Liu, Kangni and Zhang, Haotian and Tessler, Chen and Wang, Jingbo and Yuan, Ye and Cao, Jinkun and Lin, Zihui and Wang, Fengyi and others},
  journal={arXiv preprint arXiv:2407.00187},
  year={2024}
}

@inproceedings{wang2024strategy,
  title={Strategy and Skill Learning for Physics-based Table Tennis Animation},
  author={Wang, Jiashun and Hodgins, Jessica and Won, Jungdam},
  booktitle={SIGGRAPH},
  year={2024}
}

@inproceedings{zhang2024hoi,
  title={HOI-M\^{} 3: Capture Multiple Humans and Objects Interaction within Contextual Environment},
  author={Zhang, Juze and Zhang, Jingyan and Song, Zining and Shi, Zhanhe and Zhao, Chengfeng and Shi, Ye and Yu, Jingyi and Xu, Lan and Wang, Jingya},
  booktitle={CVPR},
  year={2024}
}

@article{xie2024intertrack,
  title={{InterTrack}: Tracking Human Object Interaction without Object Templates},
  author={Xie, Xianghui and Lenssen, Jan Eric and Pons-Moll, Gerard},
  journal={arXiv preprint arXiv:2408.13953},
  year={2024}
}

@article{xue2024shape,
  title={Shape Conditioned Human Motion Generation with Diffusion Model},
  author={Xue, Kebing and Seo, Hyewon},
  journal={arXiv preprint arXiv:2405.06778},
  year={2024}
}

@article{christen2024diffh2o,
  title={Diffh2o: Diffusion-based synthesis of hand-object interactions from textual descriptions},
  author={Christen, Sammy and Hampali, Shreyas and Sener, Fadime and Remelli, Edoardo and Hodan, Tomas and Sauser, Eric and Ma, Shugao and Tekin, Bugra},
  journal={arXiv preprint arXiv:2403.17827},
  year={2024}
}

@inproceedings{feng2024stratified,
  title={Stratified Avatar Generation from Sparse Observations},
  author={Feng, Han and Ma, Wenchao and Gao, Quankai and Zheng, Xianwei and Xue, Nan and Xu, Huijuan},
  booktitle={CVPR},
  year={2024}
}

@inproceedings{liu2024programmable,
  title={Programmable Motion Generation for Open-Set Motion Control Tasks},
  author={Liu, Hanchao and Zhan, Xiaohang and Huang, Shaoli and Mu, Tai-Jiang and Shan, Ying},
  booktitle={Proceedings of the IEEE/CVF Conference on Computer Vision and Pattern Recognition},
  pages={1399--1408},
  year={2024}
}

@inproceedings{song2024hoianimator,
  title={{HOIAnimator}: Generating Text-prompt Human-object Animations using Novel Perceptive Diffusion Models},
  author={Song, Wenfeng and Zhang, Xinyu and Li, Shuai and Gao, Yang and Hao, Aimin and Hou, Xia and Chen, Chenglizhao and Li, Ning and Qin, Hong},
  booktitle={CVPR},
  year={2024}
}

@inproceedings{huang2024closely,
  title={Closely Interactive Human Reconstruction with Proxemics and Physics-Guided Adaption},
  author={Huang, Buzhen and Li, Chen and Xu, Chongyang and Pan, Liang and Wang, Yangang and Lee, Gim Hee},
  booktitle={CVPR},
  year={2024}
}

@article{dai2024motionlcm,
  title={Motionlcm: Real-time controllable motion generation via latent consistency model},
  author={Dai, Wenxun and Chen, Ling-Hao and Wang, Jingbo and Liu, Jinpeng and Dai, Bo and Tang, Yansong},
  journal={arXiv preprint arXiv:2404.19759},
  year={2024}
}

@inproceedings{cen2024generating,
  title={Generating Human Motion in 3D Scenes from Text Descriptions},
  author={Cen, Zhi and Pi, Huaijin and Peng, Sida and Shen, Zehong and Yang, Minghui and Zhu, Shuai and Bao, Hujun and Zhou, Xiaowei},
  booktitle={CVPR},
  year={2024}
}

@inproceedings{li2024two,
  title={Two-Person Interaction Augmentation with Skeleton Priors},
  author={Li, Baiyi and Ho, Edmond SL and Shum, Hubert PH and Wang, He},
  booktitle={CVPR},
  year={2024}
}

@inproceedings{tang2024unified,
  title={A Unified Diffusion Framework for Scene-aware Human Motion Estimation from Sparse Signals},
  author={Tang, Jiangnan and Wang, Jingya and Ji, Kaiyang and Xu, Lan and Yu, Jingyi and Shi, Ye},
  booktitle={CVPR},
  year={2024}
}

@inproceedings{yang2024person,
  title={{Person in Place}: Generating Associative Skeleton-Guidance Maps for Human-Object Interaction Image Editing},
  author={Yang, ChangHee and Kang, ChanHee and Kong, Kyeongbo and Oh, Hanni and Kang, Suk-Ju},
  booktitle={CVPR},
  year={2024}
}

@inproceedings{diller2024futurehuman3d,
  title={{FutureHuman3D}: Forecasting Complex Long-Term 3D Human Behavior from Video Observations},
  author={Diller, Christian and Funkhouser, Thomas and Dai, Angela},
  booktitle={CVPR},
  year={2024}
}

@article{wu2024dice,
  title={{DICE}: End-to-end Deformation Capture of Hand-Face Interactions from a Single Image},
  author={Wu, Qingxuan and Dou, Zhiyang and Xu, Sirui and Shimada, Soshi and Wang, Chen and Yu, Zhengming and Liu, Yuan and Lin, Cheng and Cao, Zeyu and Komura, Taku and others},
  journal={arXiv preprint arXiv:2406.17988},
  year={2024}
}

@article{tian2024gaze,
  title={Gaze-guided hand-object interaction synthesis: Benchmark and method},
  author={Tian, Jie and Yang, Lingxiao and Ji, Ran and Ma, Yuexin and Xu, Lan and Yu, Jingyi and Shi, Ye and Wang, Jingya},
  journal={arXiv preprint arXiv:2403.16169},
  year={2024}
}

@article{zhang2024core4d,
  title={Core4d: A 4d human-object-human interaction dataset for collaborative object rearrangement},
  author={Zhang, Chengwen and Liu, Yun and Xing, Ruofan and Tang, Bingda and Yi, Li},
  journal={arXiv preprint arXiv:2406.19353},
  year={2024}
}

@article{ma2024diff,
  title={{Diff-IP2D}: Diffusion-Based Hand-Object Interaction Prediction on Egocentric Videos},
  author={Ma, Junyi and Xu, Jingyi and Chen, Xieyuanli and Wang, Hesheng},
  journal={arXiv preprint arXiv:2405.04370},
  year={2024}
}

@article{wang2024diffusion,
  title={Diffusion Models in 3D Vision: A Survey},
  author={Wang, Zhen and Li, Dongyuan and Jiang, Renhe},
  journal={arXiv preprint arXiv:2410.04738},
  year={2024}
}

@article{ma2024madiff,
  title={{MADiff}: Motion-Aware Mamba Diffusion Models for Hand Trajectory Prediction on Egocentric Videos},
  author={Ma, Junyi and Chen, Xieyuanli and Bao, Wentao and Xu, Jingyi and Wang, Hesheng},
  journal={arXiv preprint arXiv:2409.02638},
  year={2024}
}

@article{yang2024f,
  title={{F-HOI}: Toward Fine-grained Semantic-Aligned 3D Human-Object Interactions},
  author={Yang, Jie and Niu, Xuesong and Jiang, Nan and Zhang, Ruimao and Huang, Siyuan},
  journal={arXiv preprint arXiv:2407.12435},
  year={2024}
}

@article{liu2024physreaction,
  title={{PhysReaction}: Physically Plausible Real-Time Humanoid Reaction Synthesis via Forward Dynamics Guided 4D Imitation},
  author={Liu, Yunze and Chen, Changxi and Ding, Chenjing and Yi, Li},
  journal={arXiv preprint arXiv:2404.01081},
  year={2024}
}

@article{wu2024human,
  title={Human-Object Interaction from Human-Level Instructions},
  author={Wu, Zhen and Li, Jiaman and Liu, C Karen},
  journal={arXiv preprint arXiv:2406.17840},
  year={2024}
}

@article{li2024unimotion,
  title={Unimotion: Unifying 3D Human Motion Synthesis and Understanding},
  author={Li, Chuqiao and Chibane, Julian and He, Yannan and Pearl, Naama and Geiger, Andreas and Pons-Moll, Gerard},
  journal={arXiv preprint arXiv:2409.15904},
  year={2024}
}

@article{jin2024local,
  title={Local Action-Guided Motion Diffusion Model for Text-to-Motion Generation},
  author={Jin, Peng and Li, Hao and Cheng, Zesen and Li, Kehan and Yu, Runyi and Liu, Chang and Ji, Xiangyang and Yuan, Li and Chen, Jie},
  journal={arXiv preprint arXiv:2407.10528},
  year={2024}
}

@inproceedings{xusemantic,
  title={Semantic-Aware Human Object Interaction Image Generation},
  author={Xu, Zhu and Chen, Qingchao and Peng, Yuxin and Liu, Yang},
  booktitle={ICML},
  year={2024}
}

@article{zhao2024dart,
  title={{DART}: A Diffusion-Based Autoregressive Motion Model for Real-Time Text-Driven Motion Control},
  author={Zhao, Kaifeng and Li, Gen and Tang, Siyu},
  journal={arXiv preprint arXiv:2410.05260},
  year={2024}
}

@article{yang2024egochoir,
  title={{EgoChoir}: Capturing 3D Human-Object Interaction Regions from Egocentric Views},
  author={Yang, Yuhang and Zhai, Wei and Wang, Chengfeng and Yu, Chengjun and Cao, Yang and Zha, Zheng-Jun},
  journal={arXiv preprint arXiv:2405.13659},
  year={2024}
}

@article{zhong2024smoodi,
  title={{SMooDi}: Stylized Motion Diffusion Model},
  author={Zhong, Lei and Xie, Yiming and Jampani, Varun and Sun, Deqing and Jiang, Huaizu},
  journal={arXiv preprint arXiv:2407.12783},
  year={2024},
  publisher={Springer}
}

@article{zhang2024manidext,
  title={{ManiDext}: Hand-Object Manipulation Synthesis via Continuous Correspondence Embeddings and Residual-Guided Diffusion},
  author={Zhang, Jiajun and Zhang, Yuxiang and An, Liang and Li, Mengcheng and Zhang, Hongwen and Hu, Zonghai and Liu, Yebin},
  journal={arXiv preprint arXiv:2409.09300},
  year={2024}
}

@article{cao2024multi,
  title={Multi-Modal Diffusion for Hand-Object Grasp Generation},
  author={Cao, Jinkun and Liu, Jingyuan and Kitani, Kris and Zhou, Yi},
  journal={arXiv preprint arXiv:2409.04560},
  year={2024}
}

@article{daiya2024collage,
  title={{COLLAGE}: Collaborative Human-Agent Interaction Generation using Hierarchical Latent Diffusion and Language Models},
  author={Daiya, Divyanshu and Conover, Damon and Bera, Aniket},
  journal={arXiv preprint arXiv:2409.20502},
  year={2024}
}

@article{tessler2024maskedmimic,
  title={{MaskedMimic}: Unified Physics-Based Character Control Through Masked Motion Inpainting},
  author={Tessler, Chen and Guo, Yunrong and Nabati, Ofir and Chechik, Gal and Peng, Xue Bin},
  journal={arXiv preprint arXiv:2409.14393},
  year={2024}
}

@inproceedings{wu2024himo,
  title={{HIMO}: A New Benchmark for Full-Body Human Interacting with Multiple Objects},
  author={Wu, Shuwen and Liu, Yifan and Li, Lincheng and Bi, Mengxiao and Zeng, Wenjun and Yang, Xiaokang},
  year={2024},
  booktitle={ECCV},
}

@article{ma2024contact,
  title={Contact-aware Human Motion Generation from Textual Descriptions},
  author={Ma, Sihan and Cao, Qiong and Zhang, Jing and Tao, Dacheng},
  journal={arXiv preprint arXiv:2403.15709},
  year={2024}
}

@inproceedings{cui2024anyskill,
  title={{AnySkill}: Learning Open-Vocabulary Physical Skill for Interactive Agents},
  author={Cui, Jieming and Liu, Tengyu and Liu, Nian and Yang, Yaodong and Zhu, Yixin and Huang, Siyuan},
  booktitle={CVPR},
  year={2024}
}

@article{cong2024laserhuman,
  title={{LaserHuman}: Language-guided Scene-aware Human Motion Generation in Free Environment},
  author={Cong, Peishan and Dou, Ziyi WangZhiyang and Ren, Yiming and Yin, Wei and Cheng, Kai and Sun, Yujing and Long, Xiaoxiao and Zhu, Xinge and Ma, Yuexin},
  journal={arXiv preprint arXiv:2403.13307},
  year={2024}
}

@article{wu2024thor,
  title={{THOR}: Text to Human-Object Interaction Diffusion via Relation Intervention},
  author={Wu, Qianyang and Shi, Ye and Huang, Xiaoshui and Yu, Jingyi and Xu, Lan and Wang, Jingya},
  journal={arXiv preprint arXiv:2403.11208},
  year={2024}
}

@inproceedings{jiang2024scaling,
  title={Scaling Up Dynamic Human-Scene Interaction Modeling},
  author={Jiang, Nan and Zhang, Zhiyuan and Li, Hongjie and Ma, Xiaoxuan and Wang, Zan and Chen, Yixin and Liu, Tengyu and Zhu, Yixin and Huang, Siyuan},
  booktitle={CVPR},
  year={2024}
}

@article{zhang2024force,
  title={{FORCE}: Dataset and Method for Intuitive Physics Guided Human-object Interaction},
  author={Zhang, Xiaohan and Bhatnagar, Bharat Lal and Starke, Sebastian and Petrov, Ilya and Guzov, Vladimir and Dhamo, Helisa and P{\'e}rez-Pellitero, Eduardo and Pons-Moll, Gerard},
  journal={arXiv preprint arXiv:2403.11237},
  year={2024}
}

@inproceedings {casas2023smplitex,
    title = {{SMPLitex}: A Generative Model and Dataset for 3D Human Texture Estimation from Single Image},
    author = {Casas, Dan and Comino-Trinidad, Marc},
    booktitle = {BMVC},
    year = {2023}
}

@article{turk2012amazon,
  title={Amazon mechanical turk},
  author={Turk, Amazon Mechanical},
  journal={Retrieved August},
  volume={17},
  pages={2012},
  year={2012}
}

@article{wang2023physhoi,
  author    = {Wang, Yinhuai and Lin, Jing and Zeng, Ailing and Luo, Zhengyi and Zhang, Jian and Zhang, Lei},
  title     = {{PhysHOI}: Physics-Based Imitation of Dynamic Human-Object Interaction},
  journal   = {arXiv preprint arXiv:2312.04393},
  year      = {2023},
}

@inproceedings{li2024genzi,
	title={{GenZI}: Zero-Shot 3D Human-Scene Interaction Generation},
	author={Li, Lei and Dai, Angela},
	booktitle={CVPR},
	year={2024}
}

@article{wan2023tlcontrol,
  title={Tlcontrol: Trajectory and language control for human motion synthesis},
  author={Wan, Weilin and Dou, Zhiyang and Komura, Taku and Wang, Wenping and Jayaraman, Dinesh and Liu, Lingjie},
  journal={arXiv preprint arXiv:2311.17135},
  year={2023}
}

@article{zhou2023emdm,
  title={{EMDM}: Efficient Motion Diffusion Model for Fast, High-Quality Motion Generation},
  author={Zhou, Wenyang and Dou, Zhiyang and Cao, Zeyu and Liao, Zhouyingcheng and Wang, Jingbo and Wang, Wenjia and Liu, Yuan and Komura, Taku and Wang, Wenping and Liu, Lingjie},
  journal={arXiv preprint arXiv:2312.02256},
  year={2023}
}

@inproceedings{barquero2024seamless,
  title={Seamless Human Motion Composition with Blended Positional Encodings},
  author={Barquero, German and Escalera, Sergio and Palmero, Cristina},
  booktitle={CVPR},
  year={2024}
}

@article{ghosh2023remos,
  title={{ReMoS}: Reactive 3D Motion Synthesis for Two-Person Interactions},
  author={Ghosh, Anindita and Dabral, Rishabh and Golyanik, Vladislav and Theobalt, Christian and Slusallek, Philipp},
  journal={arXiv preprint arXiv:2311.17057},
  year={2023}
}

@article{wang2023intercontrol,
  title={{InterControl}: Generate Human Motion Interactions by Controlling Every Joint},
  author={Wang, Zhenzhi and Wang, Jingbo and Lin, Dahua and Dai, Bo},
  journal={arXiv preprint arXiv:2311.15864},
  year={2023}
}

@article{yazdian2023motionscript,
  title={{MotionScript}: Natural Language Descriptions for Expressive 3D Human Motions},
  author={Yazdian, Payam Jome and Liu, Eric and Cheng, Li and Lim, Angelica},
  journal={arXiv preprint arXiv:2312.12634},
  year={2023}
}

@article{liu2023interactive,
  title={Interactive Humanoid: Online Full-Body Motion Reaction Synthesis with Social Affordance Canonicalization and Forecasting},
  author={Liu, Yunze and Chen, Changxi and Yi, Li},
  journal={arXiv preprint arXiv:2312.08983},
  year={2023}
}

@article{kim2024zero,
  title={Zero-Shot Learning for the Primitives of 3D Affordance in General Objects},
  author={Kim, Hyeonwoo and Han, Sookwan and Kwon, Patrick and Joo, Hanbyul},
  journal={arXiv preprint arXiv:2401.12978},
  year={2024}
}

@article{kim2024parahome,
  title={{ParaHome}: Parameterizing Everyday Home Activities Towards 3D Generative Modeling of Human-Object Interactions},
  author={Kim, Jeonghwan and Kim, Jisoo and Na, Jeonghyeon and Joo, Hanbyul},
  journal={arXiv preprint arXiv:2401.10232},
  year={2024}
}

@inproceedings{zhao2023im,
      title={{I'M HOI}: Inertia-aware Monocular Capture of 3D Human-Object Interactions}, 
      author={Chengfeng Zhao and Juze Zhang and Jiashen Du and Ziwei Shan and Junye Wang and Jingyi Yu and Jingya Wang and Lan Xu},
      booktitle={CVPR},
      year={2024}
}

@inproceedings{yang2023lemon,
  title={{LEMON}: Learning 3D Human-Object Interaction Relation from 2D Images},
  author={Yang, Yuhang and Zhai, Wei and Luo, Hongchen and Cao, Yang and Zha, Zheng-Jun},
  booktitle={CVPR},
  year={2024}
}

@article{li2023object,
  title={Object motion guided human motion synthesis},
  author={Li, Jiaman and Wu, Jiajun and Liu, C Karen},
  journal={ACM Transactions on Graphics (TOG)},
  volume={42},
  number={6},
  pages={1--11},
  year={2023},
  publisher={ACM New York, NY, USA}
}

@article{peng2023hoi,
  title={{HOI-Diff}: Text-Driven Synthesis of 3D Human-Object Interactions using Diffusion Models},
  author={Peng, Xiaogang and Xie, Yiming and Wu, Zizhao and Jampani, Varun and Sun, Deqing and Jiang, Huaizu},
  journal={arXiv preprint arXiv:2312.06553},
  year={2023}
}

@article{li2023controllable,
  title={Controllable Human-Object Interaction Synthesis},
  author={Li, Jiaman and Clegg, Alexander and Mottaghi, Roozbeh and Wu, Jiajun and Puig, Xavier and Liu, C Karen},
  journal={arXiv preprint arXiv:2312.03913},
  year={2023}
}

@inproceedings{diller2023cg,
  title={{CG-HOI}: Contact-Guided 3D Human-Object Interaction Generation},
  author={Diller, Christian and Dai, Angela},
  booktitle={CVPR},
  year={2024}
}

@article{touvron2023llama,
  title={Llama 2: Open foundation and fine-tuned chat models},
  author={Touvron, Hugo and Martin, Louis and Stone, Kevin and Albert, Peter and Almahairi, Amjad and Babaei, Yasmine and Bashlykov, Nikolay and Batra, Soumya and Bhargava, Prajjwal and Bhosale, Shruti and others},
  journal={arXiv preprint arXiv:2307.09288},
  year={2023}
}

@inproceedings{wu2023daydreamer,
  title={Daydreamer: World models for physical robot learning},
  author={Wu, Philipp and Escontrela, Alejandro and Hafner, Danijar and Abbeel, Pieter and Goldberg, Ken},
  booktitle={CoRL},
  year={2023},
}

@inproceedings{seo2023masked,
  title={Masked world models for visual control},
  author={Seo, Younggyo and Hafner, Danijar and Liu, Hao and Liu, Fangchen and James, Stephen and Lee, Kimin and Abbeel, Pieter},
  booktitle={CoRL},
  year={2023},
}

@inproceedings{kim2020learning,
  title={Learning to simulate dynamic environments with gamegan},
  author={Kim, Seung Wook and Zhou, Yuhao and Philion, Jonah and Torralba, Antonio and Fidler, Sanja},
  booktitle={CVPR},
  year={2020}
}

@inproceedings{lin2023motionx,
  title={{Motion-X}: A Large-scale 3D Expressive Whole-body Human Motion Dataset},
  author={Lin, Jing and Zeng, Ailing and Lu, Shunlin and Cai, Yuanhao and Zhang, Ruimao and Wang, Haoqian and Zhang, Lei},
  booktitle={NeurIPS},
  year={2023}
}

@article{zhang2023artigrasp,
  title={{ArtiGrasp}: Physically Plausible Synthesis of Bi-Manual Dexterous Grasping and Articulation},
  author={Zhang, Hui and Christen, Sammy and Fan, Zicong and Zheng, Luocheng and Hwangbo, Jemin and Song, Jie and Hilliges, Otmar},
  journal={arXiv preprint arXiv:2309.03891},
  year={2023}
}

@article{braun2023physically,
  title={Physically Plausible Full-Body Hand-Object Interaction Synthesis},
  author={Braun, Jona and Christen, Sammy and Kocabas, Muhammed and Aksan, Emre and Hilliges, Otmar},
  journal={arXiv preprint arXiv:2309.07907},
  year={2023}
}

@article{xiao2023unified,
  title={Unified Human-Scene Interaction via Prompted Chain-of-Contacts},
  author={Xiao, Zeqi and Wang, Tai and Wang, Jingbo and Cao, Jinkun and Zhang, Wenwei and Dai, Bo and Lin, Dahua and Pang, Jiangmiao},
  journal={arXiv preprint arXiv:2309.07918},
  year={2023}
}

@inproceedings{brooks2023instructpix2pix,
  title={Instructpix2pix: Learning to follow image editing instructions},
  author={Brooks, Tim and Holynski, Aleksander and Efros, Alexei A},
  booktitle={CVPR},
  year={2023}
}

@article{yao2023moconvq,
  title={{MoConVQ}: Unified Physics-Based Motion Control via Scalable Discrete Representations},
  author={Yao, Heyuan and Song, Zhenhua and Zhou, Yuyang and Ao, Tenglong and Chen, Baoquan and Liu, Libin},
  journal={arXiv preprint arXiv:2310.10198},
  year={2023}
}

@inproceedings{petrovich2023tmr,
  title={{TMR}: Text-to-Motion Retrieval Using Contrastive 3D Human Motion Synthesis},
  author={Petrovich, Mathis and Black, Michael J and Varol, G{\"u}l},
  booktitle={ICCV},
  year={2023}
}

@inproceedings{guo2022tm2t,
  title={Tm2t: Stochastic and tokenized modeling for the reciprocal generation of 3d human motions and texts},
  author={Guo, Chuan and Zuo, Xinxin and Wang, Sen and Cheng, Li},
  booktitle={ECCV},
  year={2022},
}

@inproceedings{kim2023flame,
  title={Flame: Free-form language-based motion synthesis \& editing},
  author={Kim, Jihoon and Kim, Jiseob and Choi, Sungjoon},
  booktitle={AAAI},
  year={2023}
}

@inproceedings{ahuja2019language2pose,
  title={Language2pose: Natural language grounded pose forecasting},
  author={Ahuja, Chaitanya and Morency, Louis-Philippe},
  booktitle={3DV},
  year={2019},
}

@inproceedings{tevet2022motionclip,
  title={Motionclip: Exposing human motion generation to clip space},
  author={Tevet, Guy and Gordon, Brian and Hertz, Amir and Bermano, Amit H and Cohen-Or, Daniel},
  booktitle={ECCV},
  year={2022},
}

@article{zhang2023motiongpt,
  title={MotionGPT: Finetuned LLMs are General-Purpose Motion Generators},
  author={Zhang, Yaqi and Huang, Di and Liu, Bin and Tang, Shixiang and Lu, Yan and Chen, Lu and Bai, Lei and Chu, Qi and Yu, Nenghai and Ouyang, Wanli},
  journal={arXiv preprint arXiv:2306.10900},
  year={2023}
}

@inproceedings{kaufmann2020convolutional,
  title={Convolutional autoencoders for human motion infilling},
  author={Kaufmann, Manuel and Aksan, Emre and Song, Jie and Pece, Fabrizio and Ziegler, Remo and Hilliges, Otmar},
  booktitle={3DV},
  year={2020},
}

@inproceedings{rempe2023trace,
  title={Trace and Pace: Controllable Pedestrian Animation via Guided Trajectory Diffusion},
  author={Rempe, Davis and Luo, Zhengyi and Bin Peng, Xue and Yuan, Ye and Kitani, Kris and Kreis, Karsten and Fidler, Sanja and Litany, Or},
  booktitle={CVPR},
  year={2023}
}

@article{tevet2022human,
  title={Human motion diffusion model},
  author={Tevet, Guy and Raab, Sigal and Gordon, Brian and Shafir, Yonatan and Cohen-Or, Daniel and Bermano, Amit H},
  journal={arXiv preprint arXiv:2209.14916},
  year={2022}
}

@inproceedings{guo2020action2motion,
  title={Action2motion: Conditioned generation of 3d human motions},
  author={Guo, Chuan and Zuo, Xinxin and Wang, Sen and Zou, Shihao and Sun, Qingyao and Deng, Annan and Gong, Minglun and Cheng, Li},
  booktitle={ACMMM},
  year={2020}
}

@inproceedings{athanasiou2023sinc,
  title={{SINC}: Spatial Composition of 3D Human Motions for Simultaneous Action Generation},
  author={Athanasiou, Nikos and Petrovich, Mathis and Black, Michael J and Varol, G{\"u}l},
  booktitle={ICCV},
  year={2023}
}

@inproceedings{athanasiou2022teach,
  title={Teach: Temporal action composition for 3d humans},
  author={Athanasiou, Nikos and Petrovich, Mathis and Black, Michael J and Varol, G{\"u}l},
  booktitle={3DV},
  year={2022},
}

@inproceedings{zhang2023generating,
  title={Generating Human Motion From Textual Descriptions With Discrete Representations},
  author={Zhang, Jianrong and Zhang, Yangsong and Cun, Xiaodong and Zhang, Yong and Zhao, Hongwei and Lu, Hongtao and Shen, Xi and Shan, Ying},
  booktitle={CVPR},
  year={2023}
}

@article{xie2023omnicontrol,
  title={{OmniControl}: Control Any Joint at Any Time for Human Motion Generation},
  author={Xie, Yiming and Jampani, Varun and Zhong, Lei and Sun, Deqing and Jiang, Huaizu},
  journal={arXiv preprint arXiv:2310.08580},
  year={2023}
}

@inproceedings{lee2023multiact,
  title={Multiact: Long-term 3d human motion generation from multiple action labels},
  author={Lee, Taeryung and Moon, Gyeongsik and Lee, Kyoung Mu},
  booktitle={AAAI},
  year={2023}
}

@inproceedings{kong2023priority,
  title={Priority-Centric Human Motion Generation in Discrete Latent Space},
  author={Kong, Hanyang and Gong, Kehong and Lian, Dongze and Mi, Michael Bi and Wang, Xinchao},
  booktitle={ICCV},
  year={2023}
}

@article{lu2023humantomato,
  title={{HumanTOMATO}: Text-aligned Whole-body Motion Generation},
  author={Lu, Shunlin and Chen, Ling-Hao and Zeng, Ailing and Lin, Jing and Zhang, Ruimao and Zhang, Lei and Shum, Heung-Yeung},
  journal={arXiv preprint arXiv:2310.12978},
  year={2023}
}

@inproceedings{chi2023diffusionpolicy,
	title={Diffusion Policy: Visuomotor Policy Learning via Action Diffusion},
	author={Chi, Cheng and Feng, Siyuan and Du, Yilun and Xu, Zhenjia and Cousineau, Eric and Burchfiel, Benjamin and Song, Shuran},
	booktitle={RSS},
	year={2023}
}

@inproceedings{raab2023modi,
  title={{MoDi}: Unconditional Motion Synthesis from Diverse Data},
  author={Raab, Sigal and Leibovitch, Inbal and Li, Peizhuo and Aberman, Kfir and Sorkine-Hornung, Olga and Cohen-Or, Daniel},
  booktitle={CVPR},
  year={2023}
}

@inproceedings{jiang2023motiongpt,
  title={{MotionGPT}: Human Motion as a Foreign Language},
  author={Jiang, Biao and Chen, Xin and Liu, Wen and Yu, Jingyi and Yu, Gang and Chen, Tao},
  booktitle={NeurIPS},
  year={2023}
}

@misc{chatgpt,
  title = {{ChatGPT}},
  howpublished = {{\url{https://chat.openai.com/}}},
  year={2023},
  author={OpenAI}
}

@inproceedings{wei2022chain,
  title={Chain-of-thought prompting elicits reasoning in large language models},
  author={Wei, Jason and Wang, Xuezhi and Schuurmans, Dale and Bosma, Maarten and Xia, Fei and Chi, Ed and Le, Quoc V and Zhou, Denny and others},
  booktitle={NeurIPS},
  year={2022}
}

@inproceedings{zhang2023adding,
  title={Adding conditional control to text-to-image diffusion models},
  author={Zhang, Lvmin and Rao, Anyi and Agrawala, Maneesh},
  booktitle={ICCV},
  year={2023}
}

@inproceedings{brown2020language,
  title={Language models are few-shot learners},
  author={Brown, Tom and Mann, Benjamin and Ryder, Nick and Subbiah, Melanie and Kaplan, Jared D and Dhariwal, Prafulla and Neelakantan, Arvind and Shyam, Pranav and Sastry, Girish and Askell, Amanda and others},
  booktitle={NeurIPS},
  year={2020}
}

@inproceedings{mahmood2019amass,
  title={{AMASS}: Archive of motion capture as surface shapes},
  author={Mahmood, Naureen and Ghorbani, Nima and Troje, Nikolaus F and Pons-Moll, Gerard and Black, Michael J},
  booktitle={ICCV},
  year={2019}
}

@inproceedings{guo2022generating,
  title={Generating diverse and natural 3d human motions from text},
  author={Guo, Chuan and Zou, Shihao and Zuo, Xinxin and Wang, Sen and Ji, Wei and Li, Xingyu and Cheng, Li},
  booktitle={CVPR},
  year={2022}
}

@inproceedings{xu2023interdiff,
  title={{InterDiff}: Generating 3D Human-Object Interactions with Physics-Informed Diffusion},
  author={Xu, Sirui and Li, Zhengyuan and Wang, Yu-Xiong and Gui, Liang-Yan},
  booktitle={ICCV},
  year={2023}
}

@article{liang2023intergen,
  title={{InterGen}: Diffusion-based Multi-human Motion Generation under Complex Interactions},
  author={Liang, Han and Zhang, Wenqian and Li, Wenxuan and Yu, Jingyi and Xu, Lan},
  journal={arXiv preprint arXiv:2304.05684},
  year={2023}
}

@article{zhang2023roam,
      title={{ROAM}: Robust and Object-aware Motion Generation using Neural Pose Descriptors}, 
      author={Wanyue Zhang and Rishabh Dabral and Thomas Leimkühler and Vladislav Golyanik and Marc Habermann and Christian Theobalt},
      year={2023},
      journal={arXiv preprint arXiv:2308.12969},
}

@article{pan2023synthesizing,
      title={Synthesizing Physically Plausible Human Motions in 3D Scenes}, 
      author={Liang Pan and Jingbo Wang and Buzhen Huang and Junyu Zhang and Haofan Wang and Xu Tang and Yangang Wang},
      year={2023},
      journal={arXiv preprint arXiv:2308.09036},
}

@article{zhang2023tedi,
  title={{TEDi}: Temporally-Entangled Diffusion for Long-Term Motion Synthesis},
  author={Zhang, Zihan and Liu, Richard and Aberman, Kfir and Hanocka, Rana},
  journal={arXiv preprint arXiv:2307.15042},
  year={2023}
}

@article{sun2023towards,
  title={Towards Globally Consistent Stochastic Human Motion Prediction via Motion Diffusion},
  author={Sun, Jiarui and Chowdhary, Girish},
  journal={arXiv preprint arXiv:2305.12554},
  year={2023}
}

@article{xu2021d3dhoi,
  title={{D3D-HOI}: Dynamic 3D Human-Object Interactions from Videos},
  author={Xiang Xu and Hanbyul Joo and Greg Mori and Manolis Savva},
  journal={arXiv preprint arXiv:2108.08420},
  year={2021}
}

@article{xie2023hierarchical,
  title={Hierarchical Planning and Control for Box Loco-Manipulation},
  author={Xie, Zhaoming and Tseng, Jonathan and Starke, Sebastian and van de Panne, Michiel and Liu, C Karen},
  journal={arXiv preprint arXiv:2306.09532},
  year={2023}
}

@inproceedings{xie2022learning,
  title={Learning soccer juggling skills with layer-wise mixture-of-experts},
  author={Xie, Zhaoming and Starke, Sebastian and Ling, Hung Yu and van de Panne, Michiel},
  booktitle={SIGGRAPH},
  year={2022}
}

@article{yang2022learning,
  title={Learning to use chopsticks in diverse gripping styles},
  author={Yang, Zeshi and Yin, Kangkang and Liu, Libin},
  journal={ACM Transactions on Graphics (TOG)},
  volume={41},
  number={4},
  pages={1--17},
  year={2022}
}

@inproceedings{tendulkar2022flex,
    title = {{FLEX}: Full-Body Grasping Without Full-Body Grasps},
    author = {Tendulkar, Purva and Sur\'is, D\'idac and Vondrick, Carl},
    booktitle = {CVPR},
    year = {2023},
}

@inproceedings{Zhao:ICCV:2023,
   title = {Synthesizing Diverse Human Motions in 3D Indoor Scenes},
   author = {Zhao, Kaifeng and Zhang, Yan and Wang, Shaofei and Beeler, Thabo and Tang, Siyu},
   booktitle = {ICCV},
   year = {2023}
}

@inproceedings{kim2023ncho,
  title={{NCHO}: Unsupervised Learning for Neural 3D Composition of Humans and Objects}, 
  author={Taeksoo Kim and Shunsuke Saito and Hanbyul Joo},
  booktitle={ICCV},
  year={2023}
}

@inproceedings{lee2023locomotion,
  title={{Locomotion-Action-Manipulation}: Synthesizing Human-Scene Interactions in Complex 3D Environments},
  author={Lee, Jiye and Joo, Hanbyul},
  booktitle={ICCV},
  year={2023}
}

@article{hou2023compositional,
  title={Compositional 3D Human-Object Neural Animation},
  author={Hou, Zhi and Yu, Baosheng and Tao, Dacheng},
  journal={arXiv preprint arXiv:2304.14070},
  year={2023}
}

@inproceedings{zhang2022couch,
      title = {{COUCH}: Towards Controllable Human-Chair Interactions},
      author = {Zhang, Xiaohan and Bhatnagar, Bharat Lal and Starke, Sebastian and Guzov, Vladimir and Pons-Moll, Gerard},
      booktitle = {ECCV},
      year = {2022}
}

@inproceedings{zhao2022compositional,
  title={Compositional human-scene interaction synthesis with semantic control},
  author={Zhao, Kaifeng and Wang, Shaofei and Zhang, Yan and Beeler, Thabo and Tang, Siyu},
  booktitle={ECCV},
  year={2022},
}

@inproceedings{petrov2023object,
  title={Object pop-up: Can we infer 3D objects and their poses from human interactions alone?},
  author={Petrov, Ilya A and Marin, Riccardo and Chibane, Julian and Pons-Moll, Gerard},
  booktitle={CVPR},
  year={2023}
}

@inproceedings{zhou2022toch,
  title={Toch: Spatio-temporal object-to-hand correspondence for motion refinement},
  author={Zhou, Keyang and Bhatnagar, Bharat Lal and Lenssen, Jan Eric and Pons-Moll, Gerard},
  booktitle={ECCV},
  year={2022},
}

@inproceedings{wang2022reconstructing,
  title={Reconstructing Action-Conditioned Human-Object Interactions Using Commonsense Knowledge Priors},
  author={Wang, Xi and Li, Gen and Kuo, Yen-Ling and Kocabas, Muhammed and Aksan, Emre and Hilliges, Otmar},
  booktitle={3DV},
  year={2022},
}

@inproceedings{krebs2021kit,
  title={The kit bimanual manipulation dataset},
  author={Krebs, Franziska and Meixner, Andre and Patzer, Isabel and Asfour, Tamim},
  booktitle={Humanoids},
  year={2021},
}

@inproceedings{razali2023action,
  title={Action-Conditioned Generation of Bimanual Object Manipulation Sequences},
  author={Razali, Haziq and Demiris, Yiannis},
  booktitle={AAAI},
  year={2023}
}

@article{wei2023understanding,
  title={Understanding Text-driven Motion Synthesis with Keyframe Collaboration via Diffusion Models},
  author={Wei, Dong and Sun, Xiaoning and Sun, Huaijiang and Li, Bin and Hu, Shengxiang and Li, Weiqing and Lu, Jianfeng},
  journal={arXiv preprint arXiv:2305.13773},
  year={2023}
}

@inproceedings{zheng2023cams,
  title={{CAMS}: CAnonicalized Manipulation Spaces for Category-Level Functional Hand-Object Manipulation Synthesis},
  author={Zheng, Juntian and Zheng, Qingyuan and Fang, Lixing and Liu, Yun and Yi, Li},
  booktitle={CVPR},
  year={2023}
}

@inproceedings{ye2023affordance,
  title={Affordance diffusion: Synthesizing hand-object interactions},
  author={Ye, Yufei and Li, Xueting and Gupta, Abhinav and De Mello, Shalini and Birchfield, Stan and Song, Jiaming and Tulsiani, Shubham and Liu, Sifei},
  booktitle={CVPR},
  year={2023}
}

@article{li2023task,
  title={Task-Oriented Human-Object Interactions Generation with Implicit Neural Representations},
  author={Li, Quanzhou and Wang, Jingbo and Loy, Chen Change and Dai, Bo},
  journal={arXiv preprint arXiv:2303.13129},
  year={2023}
}

@inproceedings{zhang2023remodiffuse,
  title={{ReMoDiffuse}: Retrieval-Augmented Motion Diffusion Model},
  author={Zhang, Mingyuan and Guo, Xinying and Pan, Liang and Cai, Zhongang and Hong, Fangzhou and Li, Huirong and Yang, Lei and Liu, Ziwei},
  booktitle={ICCV},
  year={2023}
}

@inproceedings{dabral2023mofusion,
  title={{MoFusion}: A framework for denoising-diffusion-based motion synthesis},
  author={Dabral, Rishabh and Mughal, Muhammad Hamza and Golyanik, Vladislav and Theobalt, Christian},
  booktitle={CVPR},
  pages={9760--9770},
  year={2023}
}

@inproceedings{karunratanakul2023gmd,
title={{GMD}: Controllable Human Motion Synthesis via Guided Diffusion Models},
  author={Karunratanakul, Korrawe and Preechakul, Konpat and Suwajanakorn, Supasorn and Tang, Siyu},
  booktitle={ICCV},
  year={2023}
}

@inproceedings{chen2023executing,
  title={Executing your Commands via Motion Diffusion in Latent Space},
  author={Chen, Xin and Jiang, Biao and Liu, Wen and Huang, Zilong and Fu, Bin and Chen, Tao and Yu, Gang},
  booktitle={CVPR},
  year={2023}
}

@inproceedings{huang2023diffusion,
  title={Diffusion-based Generation, Optimization, and Planning in 3D Scenes},
  author={Huang, Siyuan and Wang, Zan and Li, Puhao and Jia, Baoxiong and Liu, Tengyu and Zhu, Yixin and Liang, Wei and Zhu, Song-Chun},
  booktitle={CVPR},
  year={2023}
}

@inproceedings{wang2022humanise,
  title={{HUMANISE}: Language-conditioned Human Motion Generation in 3D Scenes},
  author={Wang, Zan and Chen, Yixin and Liu, Tengyu and Zhu, Yixin and Liang, Wei and Huang, Siyuan},
  booktitle={NeurIPS},
  year={2022}
}

@article{kulkarni2023nifty,
  title={{NIFTY}: Neural Object Interaction Fields for Guided Human Motion Synthesis},
  author={Nilesh Kulkarni and Davis Rempe and Kyle Genova and Abhijit Kundu and Justin Johnson and David Fouhey and Leonidas Guibas},
  journal={arXiv preprint arXiv:2307.07511},
  year={2023}
}

@article{liu2018learning,
  title={Learning basketball dribbling skills using trajectory optimization and deep reinforcement learning},
  author={Liu, Libin and Hodgins, Jessica},
  journal={ACM Transactions on Graphics (TOG)},
  volume={37},
  number={4},
  pages={1--14},
  year={2018},
  publisher={ACM New York, NY, USA}
}

@inproceedings{chao2021learning,
  title={Learning to sit: Synthesizing human-chair interactions via hierarchical control},
  author={Chao, Yu-Wei and Yang, Jimei and Chen, Weifeng and Deng, Jia},
  booktitle={AAAI},
  year={2021}
}

@article{merel2020catch,
  title={Catch \& carry: reusable neural controllers for vision-guided whole-body tasks},
  author={Merel, Josh and Tunyasuvunakool, Saran and Ahuja, Arun and Tassa, Yuval and Hasenclever, Leonard and Pham, Vu and Erez, Tom and Wayne, Greg and Heess, Nicolas},
  journal={ACM Transactions on Graphics (TOG)},
  volume={39},
  number={4},
  pages={39--1},
  year={2020},
  publisher={ACM New York, NY, USA}
}

@article{starke2019neural,
  title={Neural state machine for character-scene interactions.},
  author={Starke, Sebastian and Zhang, He and Komura, Taku and Saito, Jun},
  journal={ACM Trans. Graph.},
  volume={38},
  number={6},
  pages={209--1},
  year={2019}
}

@article{starke2020local,
  title={Local motion phases for learning multi-contact character movements},
  author={Starke, Sebastian and Zhao, Yiwei and Komura, Taku and Zaman, Kazi},
  journal={ACM Transactions on Graphics (TOG)},
  volume={39},
  number={4},
  pages={54--1},
  year={2020},
  publisher={ACM New York, NY, USA}
}

@inproceedings{hassan2023synthesizing,
  title={Synthesizing Physical Character-Scene Interactions},
  author={Hassan, Mohamed and Guo, Yunrong and Wang, Tingwu and Black, Michael and Fidler, Sanja and Peng, Xue Bin},
  booktitle={SIGGRAPH},
  year={2023}
}

@inproceedings{bae2023pmp,
  title={PMP: Learning to Physically Interact with Environments using Part-wise Motion Priors},
  author={Bae, Jinseok and Won, Jungdam and Lim, Donggeun and Min, Cheol-Hui and Kim, Young Min},
  booktitle={SIGGRAPH},
  year={2023}
}

@inproceedings{chen2023humanmac,
title={{HumanMAC}: Masked Motion Completion for Human Motion Prediction},
author={Chen, Ling-Hao and Zhang, Jiawei and Li, Yewen and Pang, Yiren and Xia, Xiaobo and Liu, Tongliang},
booktitle={ICCV},
year={2023}
}

@inproceedings{
xu2023stochastic,
title={Stochastic Multi-Person 3D Motion Forecasting},
author={Sirui Xu and Yu-Xiong Wang and Liangyan Gui},
booktitle={ICLR},
year={2023},
}

@inproceedings{huang2022intercap,
    title        = {{InterCap}: {J}oint Markerless {3D} Tracking of Humans and Objects in Interaction},
    author       = {Huang, Yinghao and Taheri, Omid and Black, Michael J. and Tzionas, Dimitrios},
    booktitle    = {GCPR},
    year         = {2022}, 
}

@inproceedings{
  zhang2023neuraldome,
  title={{NeuralDome}: A Neural Modeling Pipeline on Multi-View Human-Object Interactions},
  author={Juze Zhang and Haimin Luo and Hongdi Yang and Xinru Xu and Qianyang Wu and Ye Shi and Jingyi Yu and Lan Xu and Jingya Wang},
  booktitle={CVPR},
  year={2023},
  }

@inproceedings{fan2023arctic,
  title = {{ARCTIC}: A Dataset for Dexterous Bimanual Hand-Object Manipulation},
  author = {Fan, Zicong and Taheri, Omid and Tzionas, Dimitrios and Kocabas, Muhammed and Kaufmann, Manuel and Black, Michael J. and Hilliges, Otmar},
  booktitle = {CVPR},
  year = {2023}
}

@ARTICLE {Mandery2016b,
author = {Christian Mandery and \"Omer Terlemez and Martin Do and Nikolaus Vahrenkamp and Tamim Asfour},
title = {Unifying Representations and Large-Scale Whole-Body Motion Databases for Studying Human Motion},
pages = {796--809},
volume ={32},
number ={4},
journal ={IEEE Transactions on Robotics},
year = {2016},
}

@INPROCEEDINGS {Mandery2015a,
author = {Christian Mandery and \"Omer Terlemez and Martin Do and Nikolaus Vahrenkamp and Tamim Asfour},
title = {The KIT Whole-Body Human Motion Database},
booktitle = {ICAR},
year = {2015},
}

@inproceedings{corona2020context,
  title={Context-aware human motion prediction},
  author={Corona, Enric and Pumarola, Albert and Alenya, Guillem and Moreno-Noguer, Francesc},
  booktitle={CVPR},
  year={2020}
}

@inproceedings{sohl2015deep,
  title={Deep unsupervised learning using nonequilibrium thermodynamics},
  author={Sohl-Dickstein, Jascha and Weiss, Eric and Maheswaranathan, Niru and Ganguli, Surya},
  booktitle={ICML},
  year={2015},
}

@article{song2020denoising,
  title={Denoising diffusion implicit models},
  author={Song, Jiaming and Meng, Chenlin and Ermon, Stefano},
  journal={arXiv preprint arXiv:2010.02502},
  year={2020}
}

@article{MANO,
  title = {Embodied Hands: Modeling and Capturing Hands and Bodies Together},
  author = {Romero, Javier and Tzionas, Dimitrios and Black, Michael J.},
  journal = {ACM Transactions on Graphics},
  volume = {36},
  number = {6},
  series = {245:1--245:17},
  year = {2017},
}

@inproceedings{taheri2022goal,
  title={{GOAL}: Generating 4d whole-body motion for hand-object grasping},
  author={Taheri, Omid and Choutas, Vasileios and Black, Michael J and Tzionas, Dimitrios},
  booktitle={CVPR},
  year={2022}
}

@inproceedings{wu2022saga,
  title={{SAGA}: Stochastic whole-body grasping with contact},
  author={Wu, Yan and Wang, Jiahao and Zhang, Yan and Zhang, Siwei and Hilliges, Otmar and Yu, Fisher and Tang, Siyu},
  booktitle={ECCV},
  year={2022},
}

@article{yonatan2023,
  author = {Shafir, Yonatan and Tevet, Guy and Kapon, Roy and Bermano, Amit H.},
  journal={arXiv preprint arXiv:2303.01418},
  title = {Human Motion Diffusion as a Generative Prior},
  year = {2023},
}

@inproceedings{jiang2022chairs,
  title={{CHAIRS}: Towards Full-Body Articulated Human-Object Interaction},
  author={Jiang, Nan and Liu, Tengyu and Cao, Zhexuan and Cui, Jieming and Chen, Yixin and Wang, He and Zhu, Yixin and Huang, Siyuan},
  booktitle={ICCV},
  year={2023}
}

@ARTICLE{9714029,
  author={Wan, Weilin and Yang, Lei and Liu, Lingjie and Zhang, Zhuoying and Jia, Ruixing and Choi, Yi-King and Pan, Jia and Theobalt, Christian and Komura, Taku and Wang, Wenping},
  journal={IEEE Robotics and Automation Letters}, 
  title={Learn to Predict How Humans Manipulate Large-Sized Objects From Interactive Motions}, 
  year={2022}}

@inproceedings{ho2020denoising,
  title={Denoising diffusion probabilistic models},
  author={Ho, Jonathan and Jain, Ajay and Abbeel, Pieter},
  booktitle={NeurIPS},
  year={2020}
}

@article{raab2023single,
  title={Single Motion Diffusion},
  author={Raab, Sigal and Leibovitch, Inbal and Tevet, Guy and Arar, Moab and Bermano, Amit H and Cohen-Or, Daniel},
  journal={arXiv preprint arXiv:2302.05905},
  year={2023}
}

@article{zhang2022motiondiffuse,
  title={{MotionDiffuse}: Text-driven human motion generation with diffusion model},
  author={Zhang, Mingyuan and Cai, Zhongang and Pan, Liang and Hong, Fangzhou and Guo, Xinying and Yang, Lei and Liu, Ziwei},
  journal={arXiv preprint arXiv:2208.15001},
  year={2022}
}

@inproceedings{barquero2022belfusion,
  title={{BeLFusion}: Latent Diffusion for Behavior-Driven Human Motion Prediction},
  author={Barquero, German and Escalera, Sergio and Palmero, Cristina},
  booktitle={ICCV},
  year={2023}
}

@inproceedings{bhatnagar22behave,
title = {{BEHAVE}: Dataset and Method for Tracking Human Object Interactions},
author={Bhatnagar, Bharat Lal and Xie, Xianghui and Petrov, Ilya and Sminchisescu, Cristian and Theobalt, Christian and Pons-Moll, Gerard},
booktitle = {CVPR},
year = {2022},
}

@article{loper2015smpl,
  title={{SMPL}: A skinned multi-person linear model},
  author={Loper, Matthew and Mahmood, Naureen and Romero, Javier and Pons-Moll, Gerard and Black, Michael J},
  journal={ACM transactions on graphics},
  year={2015},
}

@article{article,
author = {Kundu, Jogendra and Gor, Maharshi and Babu, R.},
year = {2019},
title = {{BiHMP-GAN}: Bidirectional 3D Human Motion Prediction GAN},
journal = {AAAI},
}

@inproceedings{mao2021generating,
  title={Generating Smooth Pose Sequences for Diverse Human Motion Prediction},
  author={Mao, Wei and Liu, Miaomiao and Salzmann, Mathieu},
  booktitle={CVPR},
  year={2021}
}

@inproceedings{yuan2020dlow,
  title={{DLow}: Diversifying latent flows for diverse human motion prediction},
  author={Yuan, Ye and Kitani, Kris},
  year = {2020},
  booktitle={ECCV},
}

@inproceedings{cao2020long,
  title={Long-term human motion prediction with scene context},
  author={Cao, Zhe and Gao, Hang and Mangalam, Karttikeya and Cai, Qi-Zhi and Vo, Minh and Malik, Jitendra},
  booktitle={ECCV},
  year={2020},
}

@inproceedings{petrovich2021action,
  title={Action-Conditioned 3D Human Motion Synthesis with Transformer VAE},
  author={Petrovich, Mathis and Black, Michael J and Varol, G{\"u}l},
  booktitle={ICCV},
  year={2021}
}

@inproceedings{petrovich22temos,
  title     = {{TEMOS}: Generating diverse human motions from textual descriptions},
  author    = {Petrovich, Mathis and Black, Michael J. and Varol, G{\"u}l},
  booktitle = {ECCV},
  year      = {2022}
}

@inproceedings{xu22stars,
  title     = {Diverse Human Motion Prediction Guided by Multi-Level Spatial-Temporal Anchors},
  author    = {Xu, Sirui and Wang, Yu-Xiong and Gui, Liang-Yan},
  booktitle = {ECCV},
  year      = {2022}
}

@inproceedings{taheri2020grab,
  title={{GRAB}: A dataset of whole-body human grasping of objects},
  author={Taheri, Omid and Ghorbani, Nima and Black, Michael J and Tzionas, Dimitrios},
  booktitle={ECCV},
  year={2020},
}

@article{ghosh2022imos,
  title={{IMoS}: Intent-Driven Full-Body Motion Synthesis for Human-Object Interactions},
  author={Ghosh, Anindita and Dabral, Rishabh and Golyanik, Vladislav and Theobalt, Christian and Slusallek, Philipp},
  journal={arXiv preprint arXiv:2212.07555},
  year={2022}
}

@inproceedings{xie2022chore,
  title={Chore: Contact, human and object reconstruction from a single rgb image},
  author={Xie, Xianghui and Bhatnagar, Bharat Lal and Pons-Moll, Gerard},
  booktitle={ECCV},
  year={2022},
}

@inproceedings{hassan2021populating,
  title={Populating 3D scenes by learning human-scene interaction},
  author={Hassan, Mohamed and Ghosh, Partha and Tesch, Joachim and Tzionas, Dimitrios and Black, Michael J},
  booktitle={CVPR},
  year={2021}
}

@inproceedings{wang2022towards,
  title={Towards diverse and natural scene-aware 3d human motion synthesis},
  author={Wang, Jingbo and Rong, Yu and Liu, Jingyuan and Yan, Sijie and Lin, Dahua and Dai, Bo},
  booktitle={CVPR},
  year={2022}
}

@inproceedings{wang2021scene,
  title={Scene-aware generative network for human motion synthesis},
  author={Wang, Jingbo and Yan, Sijie and Dai, Bo and Lin, Dahua},
  booktitle={CVPR},
  year={2021}
}

@inproceedings{wang2021synthesizing,
  title={Synthesizing long-term 3d human motion and interaction in 3d scenes},
  author={Wang, Jiashun and Xu, Huazhe and Xu, Jingwei and Liu, Sifei and Wang, Xiaolong},
  booktitle={CVPR},
  year={2021}
}

@inproceedings{zhang2020perceiving,
  title={Perceiving 3d human-object spatial arrangements from a single image in the wild},
  author={Zhang, Jason Y and Pepose, Sam and Joo, Hanbyul and Ramanan, Deva and Malik, Jitendra and Kanazawa, Angjoo},
  booktitle={ECCV},
  year={2020},
}

@inproceedings{park2019deepsdf,
  title={{DeepSDF}: Learning continuous signed distance functions for shape representation},
  author={Park, Jeong Joon and Florence, Peter and Straub, Julian and Newcombe, Richard and Lovegrove, Steven},
  booktitle={CVPR},
  year={2019}
}

@inproceedings{radford2021learning,
  title={Learning transferable visual models from natural language supervision},
  author={Radford, Alec and Kim, Jong Wook and Hallacy, Chris and Ramesh, Aditya and Goh, Gabriel and Agarwal, Sandhini and Sastry, Girish and Askell, Amanda and Mishkin, Pamela and Clark, Jack and others},
  booktitle={ICML},
  year={2021},
}

@article{van2008visualizing,
  title={Visualizing data using t-SNE.},
  author={Van der Maaten, Laurens and Hinton, Geoffrey},
  journal={Journal of machine learning research},
  volume={9},
  number={11},
  year={2008}
}

@inproceedings{rombach2022high,
  title={High-resolution image synthesis with latent diffusion models},
  author={Rombach, Robin and Blattmann, Andreas and Lorenz, Dominik and Esser, Patrick and Ommer, Bj{\"o}rn},
  booktitle={CVPR},
  year={2022}
}

@inproceedings{han2023chorus,
  title={{CHORUS}: Learning Canonicalized 3D Human-Object Spatial Relations from Unbounded Synthesized Images},
  author={Han, Sookwan and Joo, Hanbyul},
  booktitle={ICCV},
  year={2023}
}
\clearpage
\setcounter{table}{0}
\renewcommand{\thetable}{\Alph{table}}
\renewcommand*{\theHtable}{\thetable}
\setcounter{figure}{0}
\renewcommand{\thefigure}{\Alph{figure}}
\renewcommand*{\theHfigure}{\thefigure}
\setcounter{section}{0}
\renewcommand{\thesection}{\Alph{section}}
\renewcommand*{\theHsection}{\thesection}

\noindent In this Appendix, we include additional method details and experimental results: (\textbf{i}) We provide demo videos in the \href{https://sirui-xu.github.io/InterDreamer/}{website}, explained in Sec.~\ref{sec:demo}. 
(\textbf{ii}) We present additional details of interaction retrieval, world model, and optimization in Sec.~\ref{sec:method_supp}. (\textbf{iii}) We provide implementation details and additional information on the experimental setup in Sec.~\ref{sec:imple_supp}. (\textbf{iv}) We provide additional qualitative experiments in Sec.~\ref{sec:qual_supp}. (\textbf{v}) We provide some failure cases in Sec.~\ref{failure_cases_supp}. (\textbf{vi}) We discuss the potential negative societal impact in Sec.~\ref{negative impact}.

\section{Visualization Video}\label{sec:demo}
Beyond the qualitative results presented in the main paper, we include two demo videos that offer more detailed visualizations of the task, further illustrating the efficacy of our approach. These demos highlight (\textbf{i}) We conduct a qualitative comparison of our approach with existing text-to-HOI work \cite{diller2023cg,peng2023hoi} within the framework of supervised learning. Note that as our setting contains no text supervision, it is unfair to compare our work with these approaches; we include the comparison here for additional reference. We evaluate our method by directly testing our trained model on the annotated data available from their websites, specifically retrieving their generated videos for direct comparison. \emph{Remarkably, even without training on these datasets, our method generates results that demonstrate high-quality interactions.} It is even capable of synthesizing complex interactions involving \emph{dynamically-changing} contact, such as the handover and throwing of objects. 

\section{Additional Details of Methodology}\label{sec:method_supp}
\subsection{Low-Level Control} \label{sec:low_level_supp}
In this section, we provide additional details on the retrieval based on handcraft rules, which is straightforward and does not require training. We also investigate a learning-based method without relying on handcrafted designs. 

\noindent \textbf{Handcraft Interaction Retrieval.}
In Sec.~\ref{sec:low} of the main paper, we detail the construction of the interaction database and emphasize the use of body parts and object categories as keys to fetch semantically-aligned contact maps. Same as the main paper, we define a contact map as a list of \(K\) index pairs of vertices \(\{(d_h^i, d_o^i)\}_{i=1}^K\). This section delves into the methodology for outlining an optimization process to generate the object initial pose $\boldsymbol{s}_1$ given contact maps and the initial human pose $\boldsymbol{a}_1$, and choose one pose based on a predefined metric.

Let $\boldsymbol{v}_{\boldsymbol h_1}[d_o]$ denote the vertex on the surface of the object, and $\boldsymbol{v}_{\boldsymbol o_1}[d_h]$ represent the corresponding vertex on the human body surface, where $d_o$ and $d_h$ are the indices of vertices. Specifically, to optimize $\boldsymbol{s}_1$, the overall optimization objective is given by,
\begin{align}
    E_{\mathrm{opt}} = \lambda_{\mathrm{fit}}E_{\mathrm{fit}} + \lambda_{\mathrm{cont}}E_{\mathrm{cont}} + \lambda_{\mathrm{pene}}E_{\mathrm{pene}},
\end{align}
where $\lambda_{\mathrm{fit}}$, $\lambda_{\mathrm{cont}}$, and $\lambda_{\mathrm{pene}}$ are hyperparameters.

\noindent \textbf{Fitting Loss.} To project a contact map to an object pose, we minimize the L2 distance between the human vertices and the object vertices indicated by the contact map,
\begin{align}
    E_{\mathrm{fit}} = \|\boldsymbol{v}_{\boldsymbol o_1}[d_o] - \boldsymbol{v}_{\boldsymbol h_1}[d_h]\|_2.
\end{align}

\noindent \textbf{Contact Loss.} We leverage a contact loss to encourage the body part to contact the object surface in addition to the fitting loss,

\begin{align}
    E_{\mathrm{cont}} = \sum_{\tilde d_h \in \mathcal{T}}\min_{\tilde d_o}\|\boldsymbol{v}_{\boldsymbol o_1}[\tilde d_o] - \boldsymbol{v}_{\boldsymbol h_1}[\tilde d_h]\|_2,
\end{align}
where $\mathcal{T} = \{\tilde d_h | \min_{\tilde d_o}\|\boldsymbol{v}_{\boldsymbol o_1}[\tilde d_o] - \boldsymbol{v}_{\boldsymbol h_1}[\tilde d_h]\|_2 \le \epsilon\}$ includes the index of the body part that is close to the object vertex $\boldsymbol{v}_{\boldsymbol o_1}[\tilde d_o]$, where $\epsilon$ is a hyperparameter, $\tilde d_h$ and $\tilde d_o$ are vertex indices for human and object, respectively, in addition to the contact map \(\{(d_h^i, d_o^i)\}_{i=1}^K\).

\noindent \textbf{Penetration Loss.} Given the signed-distance field of the human pose $\textbf{sdf}_{\boldsymbol h_1}$, we employ a penetration loss to penalize the body-object interpenetration,

\begin{align}
    E_{\mathrm{pene}} = -\sum_{d_o}\min(\textbf{sdf}_{\boldsymbol h_1}(\boldsymbol{v}_{\boldsymbol o_1}[d_o]), 0).
\end{align}

The metric for determining the final pose selection is given by the expression $\mathds{1}(E_{\mathrm{pene}} = 0) / E_{\mathrm{cont}}$. We sample a pose from the set generated by all contact maps, with higher metrics corresponding to higher selection probability.

\noindent \textbf{Learning-Based Interaction Retrieval.}
Our interaction retrieval can also be achieved by integrating knowledge from several learning-based algorithms. Although being more complicated, the retrieval can be done without handcraft rules. Our framework can be divided into following.
(\textbf{i}) Given the text prompt $\boldsymbol{t}$ and the initial human pose $\boldsymbol{a}_1$, we synthesize corresponding images via Stable Diffusion~\cite{rombach2022high}. 
(\textbf{ii}) We follow~\cite{han2023chorus} filter out images with low quality in interaction. 
(\textbf{iii}) An off-the-shelf model LEMON~\cite{yang2023lemon} is to employed to obtain object affordance and human contact, given the generated image paired with human pose $\boldsymbol{a}_1$ and object template. The output \textbf{\(\{(l_h^i)_{i=1}^K, (l_o^i)_{i=1}^K\}\)} indicates the contact vertex indexes of human and object respectively, and the output $\boldsymbol{T}_1$ indicates the estimated object translation, which is used for initialization in the optimization.
(\textbf{iv}) To acquire the object pose, we utilize the optimization to minimize the Chamfer distance between the human vertices and the object vertices, indicated by the contact vertices obtained in the last step. 
\begin{align}
    E_{\mathrm{fit}} = \sum_{j}\min_{k}\|\boldsymbol{v}_{\boldsymbol o_1}[l_o^k] - \boldsymbol{v}_{\boldsymbol h_1}[l_h^j]\|_2.
\end{align}

\subsection{World Model}\label{sec:world_supp}

\noindent \textbf{World Model for Initial States.}
In the particular instance where the timestep $t=1$, the state vector \(\boldsymbol{s}_1\) encapsulates a single frame. Consequently, we employ two distinct models for dynamics prediction. For predictions originating from the initial state, the history motion encompasses a single timestep $H=1$. In contrast, for predictions for subsequent states, the historical interval covering \(m\) timesteps, where \(m\) denotes the frame count per segment.

\noindent \textbf{World Model for Implicit Geometry Encoding.}
The input to the world model includes the trajectories of the human vertices (represented by red small spheres in the top-right of Figure~\ref{fig:method} of the main paper), along with the vertex-to-object surface vectors. By adding the vertex-to-object surface vectors to human vertices, one can easily obtain the object vertices (shown as blue small spheres in the top-right of Figure~\ref{fig:method} of the main paper). Though the network of the world model does not receive this information directly, it can learn to combine these features to derive it when needed. 

\noindent \textbf{World Model for Novel Objects.}
The world model employs ``contact vertices'' as an input, which includes features derived from the object distance field. These features encompass the human vertex-to-object surface distance and the human vertex velocity relative to the nearest object vertex as introduced in Sec.~\ref{sec:world} of the main paper, inherently including information related to the object's shape. This encoding is consistently applied to both training objects from the BEHAVE~\cite{bhatnagar22behave} dataset and novel objects from the CHAIRS~\cite{jiang2022chairs} and OMOMO~\cite{li2023object} datasets.

\noindent \textbf{World Model for Non-Contact Objects.}
The network can process inputs without contact conditions by adopting an approach similar to ControlNet~\cite{zhang2023adding}. The network comprises two components: $\mathcal{G}$ that operates without contact vertex conditions, applicable in scenarios where no contact occurs, and $\mathcal{F}$, akin to the control components in ControlNet, which incorporates contact vertex conditions into the object trajectory when contact is present. When there is no contact, only the unconditional network is utilized. The model is aware of past object motion and thus needs to learn how human interaction affects the object’s state. This includes understanding how objects follow contact positions or normals by $\mathcal{F}$, as well as how they move without contact by $\mathcal{G}$. With the no-contact object motion data provided by BEHAVE~\cite{bhatnagar22behave}, the world model (more specifically, $\mathcal{G}$) learns to infer whether the object should free-fall based on its previous velocity or remain on the ground based on its height.

\subsection{Optimization}\label{sec:optim_supp}
We provide detailed formulations of optimization objectives, complementing Sec.~\ref{sec:optim} in the main paper.
For \textit{efficiency}, we perform optimization \textit{sparsely} only if the loss is above a threshold to improve the efficiency. 
Specifically, given the reference interaction sequence \(\{\boldsymbol{h}_i\}_{i=1}^L\) and \(\{\boldsymbol{o}_i\}_{i=1}^L\) of arbitrary length $L$, derived from previous steps, we apply gradient descent to optimize human pose sequence \(\{\boldsymbol{h}_i^\ast\}_{i=1}^L\) and object pose sequence \(\{\boldsymbol{o}_i^\ast\}_{i=1}^L\), using the loss function,
\begin{align}
    E_{\mathrm{opt}} = \lambda_{\mathrm{fit}}E_{\mathrm{fit}} + \lambda_{\mathrm{vel}}E_{\mathrm{vel}} + \lambda_{\mathrm{cont}}E_{\mathrm{cont}} + \lambda_{\mathrm{pene}}E_{\mathrm{pene}},
\end{align}
where $\lambda_{\mathrm{fit}}$, $\lambda_{\mathrm{vel}}$, $\lambda_{\mathrm{cont}}$, and $\lambda_{\mathrm{pene}}$ are hyperparameters.

\noindent \textbf{Fitting Loss.} We minimize the L1 distance between the input and the reference,
\begin{align}
    E_{\mathrm{fit}} = \sum_{i=1}^L (\|\boldsymbol{h}_i^\ast - \boldsymbol{h}_i\|_1 + \|\boldsymbol{o}_i^\ast - \boldsymbol{o}_i\|_1).
\end{align}

\noindent \textbf{Velocity Loss.} We leverage a velocity loss to smooth the interaction sequence,
\begin{align}
    E_{\mathrm{vel}} = \sum_{i=1}^{L-1} (\|\boldsymbol{h}_{i+1}^\ast - \boldsymbol{h}_i^\ast\|_1 + \|\boldsymbol{o}_{i+1}^\ast - \boldsymbol{o}_i^\ast\|_1).
\end{align}

\noindent \textbf{Contact loss.} We leverage a contact loss to encourage the body part to contact the object surface, if they are close to each other in the initial interaction, 
\begin{align}
    E_{\mathrm{cont}} = \sum_{i=1}^{L}\sum_{d_h \in \mathcal{T}_i}\min_{d_o}\|\boldsymbol{v}_{\boldsymbol o_i^\ast}[d_o] - \boldsymbol{v}_{\boldsymbol h_i^\ast}[d_h]\|_2,
\end{align}
where $\boldsymbol{v}_{\boldsymbol h_i^\ast}[d_h]$ denotes the vertex on the human body surface, and $\boldsymbol{v}_{\boldsymbol o_i^\ast}[d_o]$ represents the corresponding vertex on the surface of the object. And $\mathcal{T}_i = \{d_h | \min_{d_o}\|\boldsymbol{v}_{\boldsymbol o_i}[d_o] - \boldsymbol{v}_{\boldsymbol h_i}[d_h]\|_2 \le \epsilon\}$ includes the index of reference human vertex $\boldsymbol{v}_{\boldsymbol h_i}[d_h]$ that is close to the reference object vertex $\boldsymbol{v}_{\boldsymbol o_i}[d_o]$, where $\epsilon$ is a hyperparameter, $d_h$ and $d_o$ are vertex indices for human and object, respectively.

\noindent \textbf{Penetration Loss.} Given the signed-distance field of the human pose $\textbf{sdf}_{\boldsymbol h_i^\ast}$, we employ a penetration loss to penalize the body-object interpenetration,
\begin{align}
    E_{\mathrm{pene}} = -\sum_{i=1}^{L}\sum_{d_o}\min(\textbf{sdf}_{\boldsymbol h_i^\ast}(\boldsymbol{v}_{\boldsymbol o_i^\ast}[d_o]), 0).
\end{align}

\section{Additional Details of Experimental Setup}\label{sec:imple_supp}
\begin{figure}
    \centering
    \includegraphics[width=\columnwidth]{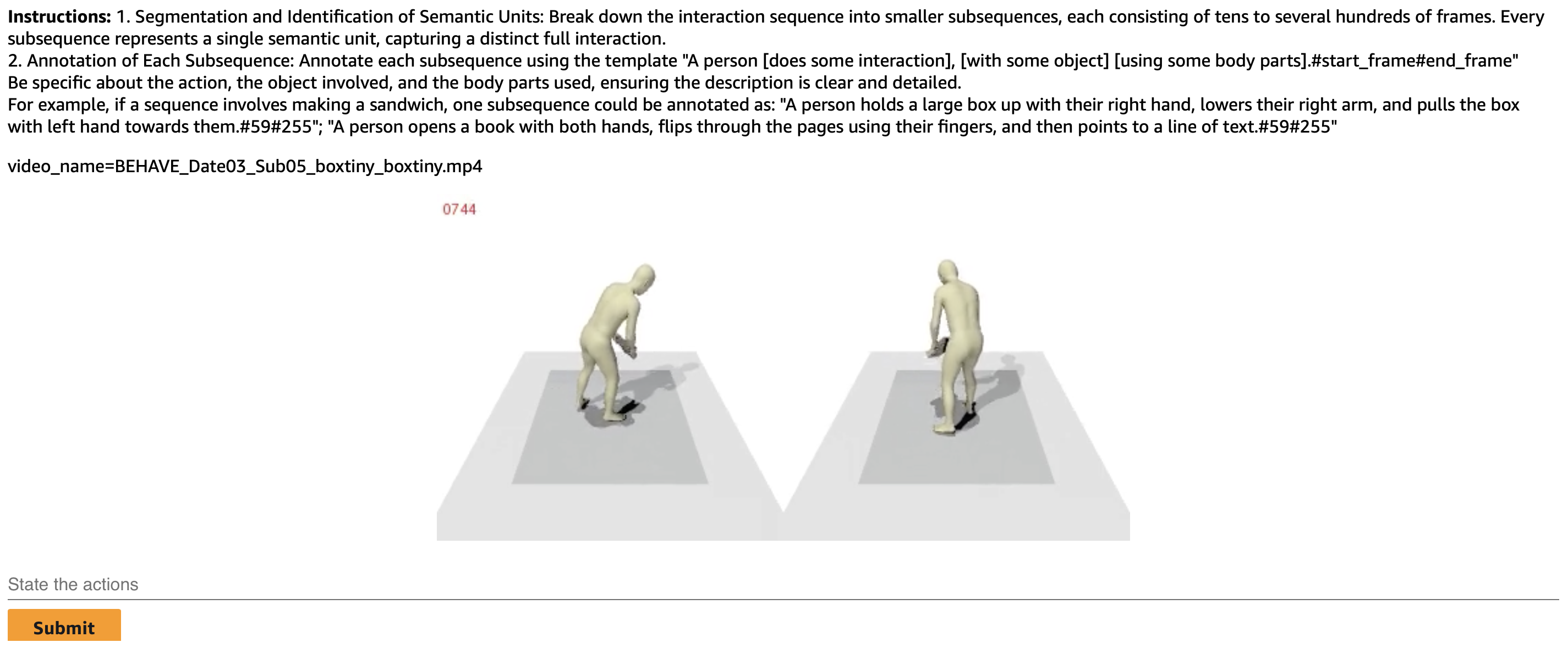}
    \caption{We use Amazon Mechanical Turk~\cite{turk2012amazon} to build an annotation platform. We provide instructions to guide the annotator to split a long sequence into several short sub-sequences with their start and end frames, and then annotate each sub-sequence. We inform annotators that our collected data are used for text-motion generation when they accept the job.
    }
    \label{fig:annot}
\end{figure}
\begin{figure}
    \centering
   \includegraphics[width=\textwidth]{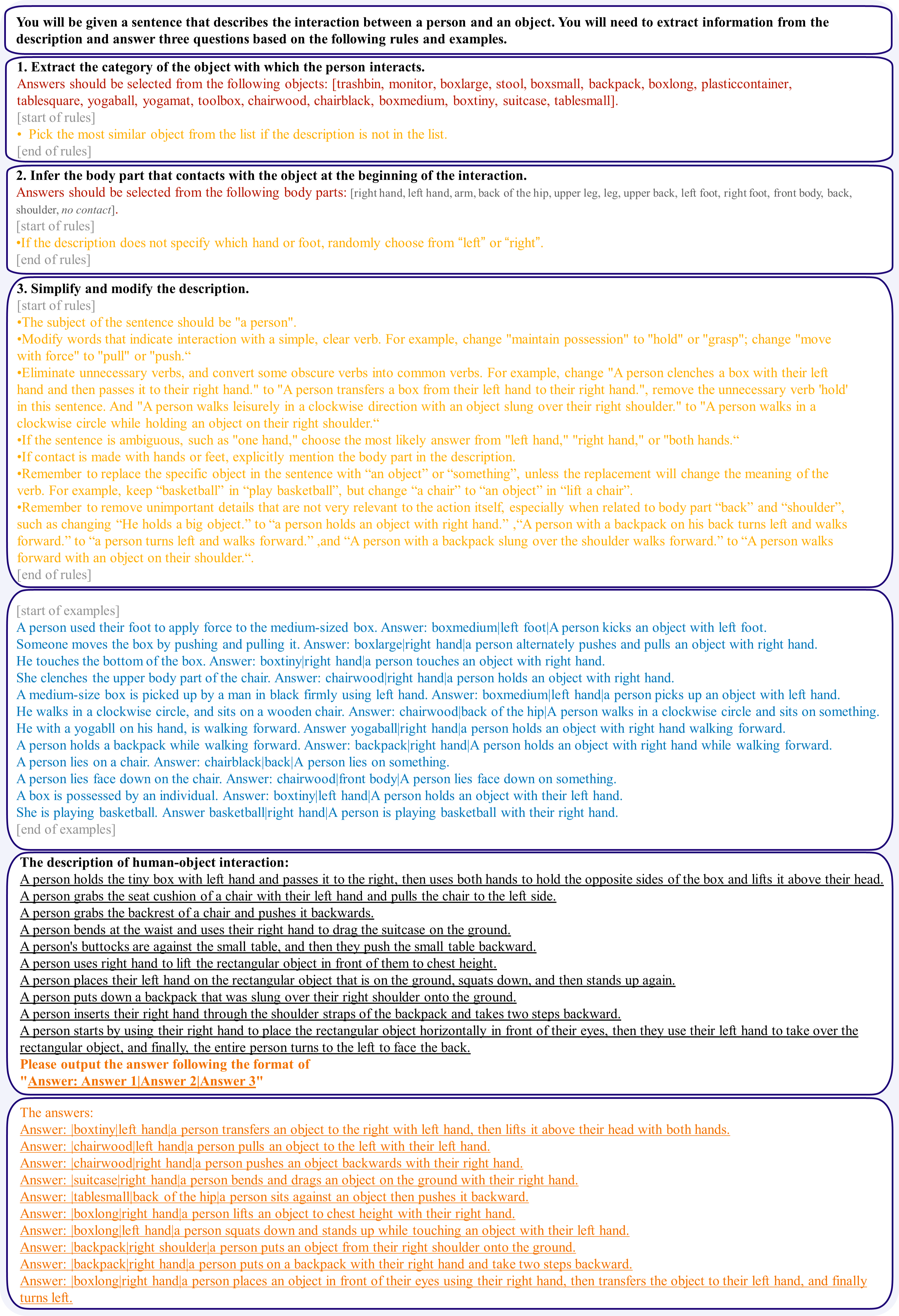}
    \caption{Full log of our high-level planning.}
    \label{fig:protocol}

\end{figure}

\begin{figure}
    \centering
    \includegraphics[width=\columnwidth]{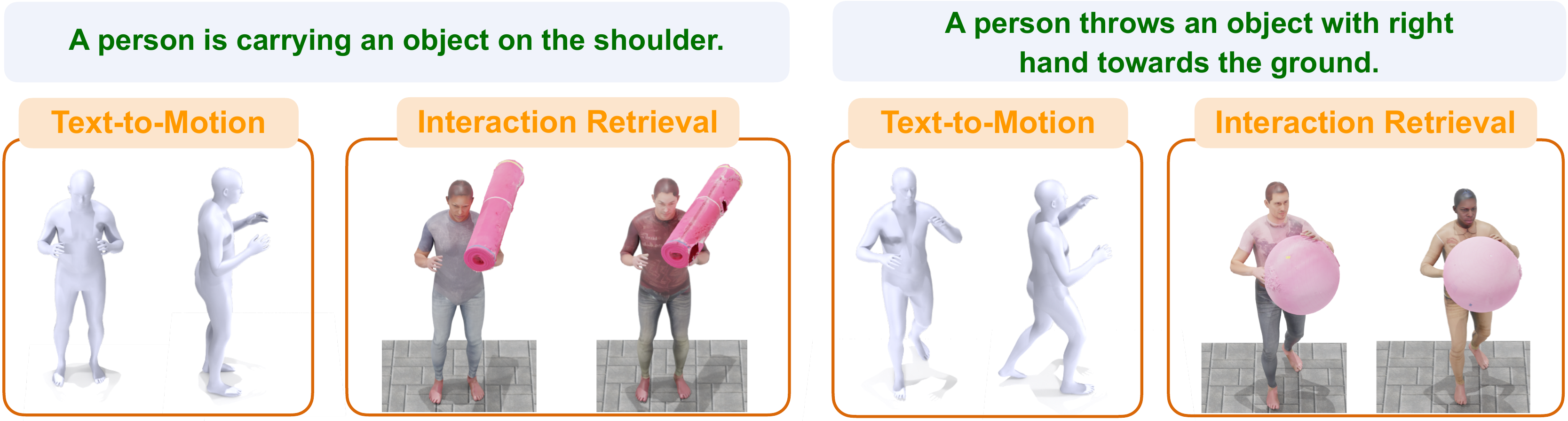}
    \caption{\textbf{Qualitative results} from the interaction retrieval. We demonstrate that our learning-based interaction retrieval can extract diverse and realistic interactions.
    }
    \label{fig:retrie}
\end{figure}
\noindent \textbf{Datasets.} We include a screenshot of our annotation platform in Figure~\ref{fig:annot}. Our annotations are further diversified by GPT-4~\cite{chatgpt}. The prompt used for this purpose is: \texttt{I'm going to give you a description, and I would like to have three rewritten sentences with varying degrees of complexity, following the example: ``...'' The input text is ``...'' Please give me three texts that vary in complexity but keep the meaning of the sentence the same.} This results in \textbf{(i)} \textit{less complexity}: someone holds a backpack and steps left; \textbf{(ii)} \textit{middle complexity}: a person holds a backpack in front of them with both hands and takes a step to the left; \textbf{(iii)} \textit{more complexity}: with both hands, a person clutches a heavy backpack firmly and brings it close to their body, then steps to the left with their left leg.

\noindent \textbf{Metrics.} In Sec.~\ref{sec:exp_set}, we introduce the metrics employed in this paper. This section details the formula for the proposed metric CMD. The formulations for other metrics are available in the existing literature~\cite{xu2023interdiff, guo2022generating}. CMD quantifies the discrepancy between the contact maps of ground truth interactions and those synthesized one. In this context, a contact map is characterized by the proportion of time \(\{\boldsymbol{p}_{i}\}_{i=1}^P\) each body part maintains active contact. Here, \(\boldsymbol{p}_{i}\) denotes the percentage of time during which the body part \(i\) is less than a threshold distance from the object. And the metric is defined as,
\begin{align}
    \mathrm{CMD} = \frac{1}{P}\sum_{i=1}^P\|\boldsymbol{p}_{i} - \boldsymbol{p}_{i}^{\mathrm{GT}}\|_1,
\end{align}
where $\boldsymbol{p}_{i}^{\mathrm{GT}}$ is from the ground truth contact map, $P$ is the number of body parts defined in SMPL~\cite{loper2015smpl}, and we set the distance threshold as $0.03$ m.

\noindent\textbf{Implementation Details.} The segment in the MDP contains \(m=4\) frames. The dynamics model, which includes 2 dynamics blocks as described in the main paper, is trained on the BEHAVE training set~\cite{bhatnagar22behave}, with a batch size of 32, a latent dimension of 64, and for 500 epochs. For rollout after the initial step $t > 1$, our dynamics model is trained to predict over a longer timeframe (\( F = 3 \times m = 12 \)), exceeding the past motion duration (\( H = m = 4 \)). For the initial step $t = 1$, we train a separate dynamics model to forecast a duration of \( F = 15 \) given the past motion over \( H = 1 \) frame, consistent with Sec.~\ref{sec:world_supp}. The optimization process is conducted over $300$ epochs, utilizing a learning rate of $0.01$. The dynamic model is trained on an NVIDIA A40 GPU for a day. Our full log for high-level planning is presented in Figure~\ref{fig:protocol}.

\section{Additional Qualtitative Results}\label{sec:qual_supp}

\noindent\textbf{Interaction Retrieval.}
We here visualize the intermediate retrieval results. Figure~\ref{fig:retrie} depicts the results from learning-based retrieval, resulting in a diverse set of interactions that are both high-quality and semantically aligned.

\begin{figure*}
    \centering
    \includegraphics[width=\textwidth]{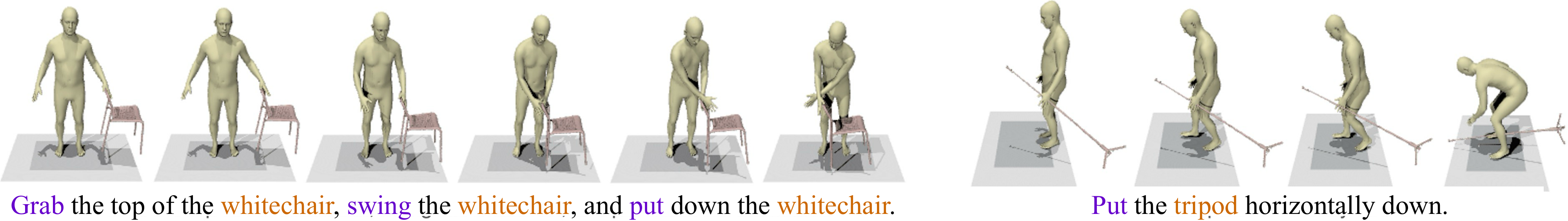}
     \caption{\textbf{Qualitative results} on the OMOMO~\cite{li2023object} dataset. Our method generalizes well on the OMOMO objects and annotations unseen in training. Frames are separately visualized. Here, our synergized models are GPT-4~\cite{chatgpt} and MotionGPT~\cite{jiang2023motiongpt}.} 
    \label{fig:omomo}
\end{figure*}

\noindent\textbf{Qualitative Experiments on OMOMO dataset.}
Figure~\ref{fig:omomo} exemplifies our method that is able to generalize effectively to the OMOMO~\cite{li2023object} dataset, despite our dynamics model not being trained on its object geometry or annotations.

\section{Failure Cases}
\label{failure_cases_supp}

We present failure cases of our method in our \href{https://sirui-xu.github.io/InterDreamer/}{website}, consisting of (i) inconsistency of interaction with textual description, (ii) inconsistency of human actions with textual descriptions, and (iii) wrong object category inferred by LLM.

\section{Potential Negative Societal Impact} \label{negative impact}

Some potential negative societal impacts include: (\textbf{i}) Our approach can synthesize realistic human motion interacting with objects, which could be misused to create deceptive or harmful content, such as portraying individuals in false situations. This could contribute to the spread of misinformation.
(\textbf{ii}) Our method evaluates real behavioral information, raising potential privacy concerns. Although our model utilizes a processed representation (SMPL~\cite{loper2015smpl}) of human motion that retains minimal identifying details -- unlike raw data or images -- its ability to simulate human-object interactions could still be exploited for unauthorized surveillance or behavioral analysis. For instance, with photorealistic textures, it might be used to model and generate personal habits or movements without consent, posing risks of privacy violations. However, the use of a processed representation can be positively viewed as a privacy-enhancing feature, as it minimizes the exposure of personally identifiable details.

\end{document}